# MeanCut: A Greedy-Optimized Graph Clustering via Path-based Similarity and Degree Descent Criterion

Dehua Peng, Zhipeng Gui and Huayi Wu

**Abstract**—As the most typical graph clustering method, spectral clustering is popular and attractive due to the remarkable performance, easy implementation, and strong adaptability. Classical spectral clustering measures the edge weights of graph using pairwise Euclidean-based metric, and solves the optimal graph partition by relaxing the constraints of indicator matrix and performing Laplacian decomposition. However, Euclidean-based similarity might cause skew graph cuts when handling non-spherical data distributions, and the relaxation strategy introduces information loss. Meanwhile, spectral clustering requires specifying the number of clusters, which is hard to determine without enough prior knowledge. In this work, we leverage the path-based similarity to enhance intra-cluster associations, and propose MeanCut as the objective function and greedily optimize it in degree descending order for a nondestructive graph partition. This algorithm enables the identification of arbitrary shaped clusters and is robust to noise. To reduce the computational complexity of similarity calculation, we transform optimal path search into generating the maximum spanning tree (MST), and develop a fast MST (FastMST) algorithm to further improve its time-efficiency. Moreover, we define a density gradient factor (DGF) for separating the weakly connected clusters. The validity of our algorithm is demonstrated by testifying on real-world benchmarks and application of face recognition. The source code of MeanCut is available at https://github.com/ZPGuiGroupWhu/MeanCut-Clustering.

**Index Terms**—Graph partition, spectral clustering, degree of vertex, maximum spanning tree.

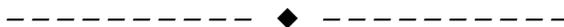

## 1 Introduction

CLUSTRING is a powerful machine learning method to explore the similar patterns and has been widely used in many fields, e.g., computer science, geoscience, biology, economics and so on [1]-[3]. To cope with different challenges in cluster analysis, a multitude of clustering algorithms have been proposed and improved. They can be categorized into partition-based, density-based, grid-based, prototype-based, direction-based, hierarchical, and graph clustering [4]. Among these algorithms, graph clustering conforms the primitive idea of clustering that is maximizing the intra-cluster similarity and inter-cluster difference. Specifically, it establishes associations between points in the form of graph, and partitions the graph into multiple disconnected subgraphs based on decomposition rules. Each separated subgraph corresponds a cluster.

The essentials of graph clustering include the construction of graph, measurement of edge weights, and optimization of graph cut function. In the raw data, each point is independent without any associations. Graph is constructed to connect them based on the distance proximity. ε-neighborhood, K-Nearest Neighbor (KNN) and fully connected graphs are three commonly used ways to link points into a graph. ε-neighborhood graph builds associations in the range of a fixed distance ε. The distance threshold has significant impact on the graph that low-density points tend to be isolated without connections if the global ε is not large enough. KNN-based graph requires the two vertices of each generated edge to be mutual KNNs. Points in sparse clusters can be connected, but the clusters with heterogeneous density are easily to be split to multiple unconnected subgraphs. Fully connected graph generates edges between any pair of points, but dense edges also increase the time consumption of graph partition. For measuring the edge weights, the similarities of point pairs, e.g., the Euclidean distance, are calculated, and kernel functions are leveraged to normalize the edge weights between 0 and 1. Radial basis function (RBF) transforms the Euclidean distance from a linear to a convergent nonlinear space. It includes Gaussian, Exponential, Laplacian kernels and so on that have received extensive use. The optimization of graph cut function is the most challenging and crucial issue in graph clustering. Theoretically, solving the optimal graph partition is a NP-hard problem, which has exponential time complexity to traverse all the possible solutions.

To make the optimal graph partition easy to solve, spectral clustering relaxes the hard constraints of objective function (i.e., graph cut function), thereby transforming a NP-hard problem to the spectral decomposition of Laplacian matrix. The procedure of classical spectral clustering is a two-stage process, e.g., embedding and discretization [5]. It constructs a graph firstly, and calculates the weight, degree and Laplacian matrices. Laplacian matrix equals to the degree matrix minus weight (similarity) matrix, which is symmetric and positive semi-definite. Then, the spectral embedding is performed by solving the first $k$ ($k$ denotes the number of clusters) eigenvalues in an ascending sort and corresponding eigenvectors of Laplacian (or normalized Laplacian). In the process of discretization, K-means

---

- *D. Peng and H. Wu are with the State Key Laboratory of Information Engineering in Surveying, Mapping and Remote Sensing, Wuhan University, Wuhan 430079, China, and also with Collaborative Innovation Center of Geospatial Technology, Wuhan University, Wuhan 430079, China. E-mail: {pengdh, wuhuayi}@ whu.edu.cn.*
- *Z. Gui is with the School of Remote Sensing and Information Engineering, Wuhan University, Wuhan 430079, China, and also with Collaborative Innovation Center of Geospatial Technology, Wuhan University, Wuhan 430079, China. E-mail: zhipeng.gui@whu.edu.cn.*

is used to cluster the matrix composed of the $k$ eigenvectors and obtain the clustering results.

In spectral clustering, a variety of graph cut functions have been proposed, such as Minimum cut (Mincut) [6], Ratio cut (Rcut) [7], Normalized cut (Ncut) [8], min-max cut [9] and so on [10]. Mincut is aimed at minimizing the sum of edge weights between different clusters. However, Mincut tends to separate individual isolated points from the graph in many cases, which conflicts with the common observation that a cluster should be a reasonably large group of points. To tackle this issue, Rcut considers the number of points within clusters on the basis of Mincut, while Ncut keeps the intra-cluster edge weights at a high level when minimizing the inter-cluster associations. However, unbalanced cuts still occur when handling non-spherical clusters with heterogeneous densities. Min-max cut simultaneously minimizes the similarities between different subgraphs and maximizes that within each subgraph, but it requires the number of clusters and is sensitive to noise points.

Although the abovementioned graph cuts have been widely used, there are still some problems need to be addressed: 1) Relaxation of the constraint conditions and using the auxiliary clustering algorithms (e.g., K-means) would bring information loss. Indicator matrix utilizes binary discrete elements to store the subordinate relations between the points and clusters. In order to simplify the optimization of graph partition, indicator matrix are relaxed into be a continuous space by spectral embedding, which makes it unable to indicate the subordinate relations. Auxiliary clustering algorithms are exploited to convert the solved indicator matrix into discrete clustering instead, but they cannot restore the indicative information nondestructively. 2) Arbitrary shapes and heterogeneous densities of data distribution may cause skew cuts [11]. Classical spectral clustering utilizes Euclidean-based distance to measure the pairwise similarities. In this situation, a large number of point pairs belonging to different clusters have higher similarities than intra-cluster ones, especially those in none-spherical clusters with proximate distance. Consequently, similarity matrix does not present obvious blocky structures, producing adverse effect on graph partition. 3) The matrix decomposition requires the accurate number of clusters. It is a sensitive parameter and affects the clustering results significantly. Usually, the number of clusters is hard to determine, since the prior knowledge of distribution is lacking for unexplored data. 4) Classical spectral clustering is not robust to the noise [12].

Actually, these problems have received attention. To avoid the information loss, the trace minimization problem of spectral clustering was transformed into the objective of kernel K-means, and a novel multilevel graph partitioning method was proposed to optimize the objective function [13]. Besides, agglomerative hierarchical clustering was exploited to optimize Ncut greedily, but the computational complexity is high when the graph is complete and the dendrogram is unbalanced [14]. In terms of the similarity measurement, a path-based similarity (i.e., connectivity distance) was developed [15]-[17], which inherits the characteristics of Euclidean distance, i.e., symmetric, non-negative and reflexivity. It facilitates to identify the clusters of non-spherical shapes and manifold structures, but the calculation process is expensive, and is hard to handle weakly connected clusters. Meanwhile, determining the number of clusters matters. The optimal number of clusters can be determined using Silhouette index [18]. However, it produces a series of invalid clustering results and introduces extra time cost. Considering that not all eigenvectors are equally informative and useful, a relevance learning method was proposed to remove the redundant and irrelevant eigenvectors [12]. Nonetheless, it is sensitive to parameter initialization especially on noisy and sparse data. Noise would corrupt the similarity matrix and influence the global eign-decomposition of matrix. The interference of noise can be reduced by introducing a regularization term to punish the objective function and achieving row sparsity, yet the approach is super sensitive to the parameters [19].

In this paper, we propose a novel graph clustering algorithm upon path-based similarity and degree descent criterion. In order to preserve the complete information of indicator matrix, we define a graph cut function called MeanCut and conduct greedy optimization instead of spectral decomposition. The path-based similarity enhances the intra-cluster associations and makes the separability between clusters more distinct. Degree descent is proved to be the best order to minimize MeanCut and agglomerate points into clusters. For the sake of time efficiency, we propose a FastMST algorithm with a time complexity of $O(nlogn)$ for similarity calculation. In general, MeanCut clustering dispenses with prior knowledge of the number of clusters, and is competent for noise elimination. Moreover, a density gradient factor (DGF) is designed to remove the points at the junction or boundary areas of clusters, enabling the separation of weakly connected clusters.

This paper is organized as follows: Section 2 introduces the background of typical graph cut functions. Section 3 proposes the fundamentals of the parameter-free MeanCut algorithm. Section 4 presents improvement strategies for time efficiency and clustering accuracy. Section 5 evaluates the effectiveness on real-world and synthetic datasets, and testifies our algorithm on face recognition. Section 6 draws the conclusion.

## 2 BACKGROUND OF SPECTRAL CLUSTERING

This section presents the basic concepts and typical graph cuts in spectral clustering. Rayleigh quotient is introduced to interpret the spectral theory. Table 1 defines the terms and notations used throughout the paper.

### 2.1 Basic Concepts

Given $n$ data points $\mathbf{V} = \{v_1, v_2, \dots, v_n\} \in \mathbb{R}^n$, a graph can be constructed and denoted as $\mathbf{G} = (\mathbf{V}, \mathbf{E}, \mathbf{W})$. Here, we construct a fully connected graph in which any two points are linked with an edge. Based on Euclidean distance,

TABLE 1
Overview of the notations.

| Notation | Description |
| --- | --- |
| $n$ | Number of the vertices |
| $k$ | Number of the expected clusters |
| $K$ | K-Nearest Neighbor |
| $v_i$ | Vertex $i$ in the graph |
| $w_{ij}$ | Similarity (edge weight) between $v_i$ and $v_j$ |
| $\hat{w}_{ij}$ | Path-based similarity between $v_i$ and $v_j$ |
| $d_i$ | Degree of vertex $v_i$ |
| $R$ | Rayleigh quotient |
| $OP_{ij}$ | The optimal path between $v_i$ and $v_j$ |
| $P_{ij}$ | Paths between $v_i$ and $v_j$ |
| $G$ | The constructed graph |
| $V \in \mathbb{R}^n$ | The set of vertices of graph $G$ |
| $E \in \mathbb{R}^{n \times n}$ | The set of edges of graph $G$ |
| $W \in \mathbb{R}^{n \times n}$ | Similarity (weight) matrix |
| $D \in \mathbb{R}^{n \times n}$ | Degree matrix |
| $L \in \mathbb{R}^{n \times n}$ | Laplacian matrix |
| $X \in \mathbb{R}^{n \times k}$ | Indicator matrix |
| $x_1, x_2, ..., x_k$ | Indicator vectors of the $k$ clusters |
| $\lambda_1, \lambda_2, ..., \lambda_k$ | The first $k$ eigenvalues |
| $\mathbb{v}_1, \mathbb{v}_2, ..., \mathbb{v}_k$ | The first $k$ eigenvectors |
| $C_1, C_2, ..., C_k$ | The $k$ connected components (clusters) |

Gaussian kernel is commonly used to measure the similarity between vertices $v_i$ and $v_j$ as

$$w_{ij} = \exp\left(-\frac{\|v_i - v_j\|^2}{2\sigma^2}\right) \quad (1)$$

In spectral clustering, degree of a vertex is defined as the weight sum of all edges incident to vertex $v_i$

$$d_i = \sum_{j=1}^{n} w_{ij} \quad (2)$$

Thus, the degree matrix can be denoted as a diagonal matrix $D = \text{diag}(d_1, d_2, ..., d_n) \in \mathbb{R}^{n \times n}$. The Laplacian matrix is defined as

$$L = D - W \quad (3)$$

$L \in \mathbb{R}^{n \times n}$ is a symmetric and positive semi-definite matrix. Given a vector $f^T = [f_1, f_2, ..., f_n] \in \mathbb{R}^n$, we have $f^T L f = \frac{1}{2} \sum_{i,j=1}^{n} w_{ij}(f_i - f_j)^2$. Laplacian matrix has $n$ non-negative and real-valued eigenvalues, where the smallest eigenvalue is zero, and the corresponding eigenvector is a $\mathbf{1}$ vector.

Spectral clustering aims to cut an undirected graph $G$ into $k$ disconnected subgraphs $\{C_1, C_2, ..., C_k\}$, which form a partition if any two subgraphs $C_i \cap C_j = \emptyset$ and $C_1 \cup C_2 \cup ... \cup C_k = V$. Each subgraph is called a connected component. There are no connections between the vertices in $C$ and its complement set $\overline{C}$. We define the graph cut between two connected components as the weight sum of inter-cluster edges

$$cut(C_i, C_j) = \sum_{v_s \in C_i, v_t \in C_j} w_{st} \quad (4)$$

For the global $k$ connected components, the $k$-way graph cut can be denoted as

$$cut(C_1, C_2, ..., C_k) = \sum_{i=1}^{k} cut(C_i, \overline{C_i}) \quad (5)$$

Mincut is defined as the minimum of the $k$-way graph cut in Eq. (5). To solve it, we introduce the indicator matrix $X = [x_1, x_2, ..., x_k] \in \mathbb{R}^{n \times k}$, which is composed of $k$ orthogonal indicator vectors $x_i^T = [x_{i1}, x_{i2}, ..., x_{in}] \in \mathbb{R}^n$

$$x_{ij} = \begin{cases} 1 & \text{if } v_j \in C_i \\ 0 & \text{if } v_j \notin C_i \end{cases} \quad (6)$$

Using the indicator vectors and Laplacian matrix, the graph cut can be transformed as

$$\begin{aligned} x_i^T L x_i &= \frac{1}{2} \sum_{s,t=1}^{n} w_{st}(x_{is} - x_{it})^2 \\ &= \frac{1}{2} \Big( \sum_{v_s \in C_i, v_t \notin C_i} w_{st}(1-0)^2 + \sum_{v_s \notin C_i, v_t \in C_i} w_{st}(0-1)^2 \Big) \\ &= \frac{1}{2} \Big( \sum_{v_s \in C_i, v_t \notin C_i} w_{st} + \sum_{v_s \notin C_i, v_t \in C_i} w_{st} \Big) \\ &= \frac{1}{2} \big( cut(C_i, \overline{C_i}) + cut(\overline{C_i}, C_i) \big) = cut(C_i, \overline{C_i}) \quad (7) \end{aligned}$$

Thus, the Mincut can be represented as

$$Mincut(X) = \sum_{i=1}^{k} (X^T L X)_{ii} = tr(X^T L X) \quad (8)$$

However, Mincut does not consider the cluster sizes (i.e., the number of points in cluster), so the low-density or isolated vertices tends to be identified as small clusters.

## 2.2 Rcut and Ncut
To avoid skew cuts in Mincut, Rcut takes account of the cluster size and introduces the number of vertices of each subgraph. It is defined as

$$Rcut(X) = \sum_{i=1}^{k} \frac{cut(C_i, \overline{C_i})}{|C_i|} = \sum_{i=1}^{k} \frac{x_i^T L x_i}{x_i^T x_i} \quad (9)$$

We transform the indicator vectors to the orthonormal vectors $\tilde{x}_i = x_i (x_i^T x_i)^{-1/2}$ that will form a normalized indicator matrix $\tilde{X}$. Then, the objective function of Rcut can be represented as

$$\min Rcut(X) = \min tr\left(\tilde{X}^T L \tilde{X}\right) s.t. \tilde{X}^T \tilde{X} = I \quad (10)$$

To enhance the intra-cluster similarities, Ncut introduces the sum of the degree of all the points in each subgraph, which can be denoted as

$$Ncut(X) = \sum_{i=1}^{k} \frac{cut(C_i, \overline{C_i})}{\sum_{v_j \in C_i} d_j} = \sum_{i=1}^{k} \frac{x_i^T L x_i}{x_i^T D x_i} \quad (11)$$

Let $y_i = (x_i^T D x_i)^{-1/2} D^{1/2} x_i$ and $Y = (X^T D X)^{-1/2} D^{1/2} X$,

then **Y** holds

$$\mathbf{Y}^T\mathbf{Y} = \left((\mathbf{X}^T\mathbf{D}\mathbf{X})^{-1/2}\mathbf{D}^{1/2}\mathbf{X}\right)^T\left((\mathbf{X}^T\mathbf{D}\mathbf{X})^{-1/2}\mathbf{D}^{1/2}\mathbf{X}\right) = \mathbf{I} \quad (12)$$

Under the constraint of Eq. (12), the Ncut problem can be formulated as

$$\min Ncut(\mathbf{X}) = \min tr\left(\frac{\mathbf{X}^T\mathbf{L}\mathbf{X}}{\mathbf{X}^T\mathbf{D}\mathbf{X}}\right)$$

$$= \min tr\left(\frac{((\mathbf{X}^T\mathbf{D}\mathbf{X})^{1/2}\,\mathbf{Y}^T\mathbf{D}^{-1/2})\mathbf{L}((\mathbf{X}^T\mathbf{D}\mathbf{X})^{1/2}\mathbf{D}^{-1/2}\mathbf{Y})}{\mathbf{X}^T\mathbf{D}\mathbf{X}}\right)$$

$$= \min tr(\mathbf{Y}^T\mathbf{D}^{-1/2}\mathbf{L}\mathbf{D}^{-1/2}\mathbf{Y})\ s.t.\ \mathbf{Y}^T\mathbf{Y} = \mathbf{I} \quad (13)$$

It is a NP-hard problem to solve the optimal indicator matrices of Rcut and Ncut directly. The theory of Rayleigh quotient provides a relaxation strategy to find the optimal solution of Rcut and Ncut.

### 2.3 Rayleigh Quotient

Given a non-zero vector $\boldsymbol{f} \in \mathbb{R}^n$ and a real symmetric matrix $\mathbf{A} \in \mathbb{R}^{n \times n}$ that satisfies $\mathbf{A}^T = \mathbf{A}$, Rayleigh quotient can be defined as

$$R(\mathbf{A}, \boldsymbol{f}) = \frac{\boldsymbol{f}^T \mathbf{A} \boldsymbol{f}}{\boldsymbol{f}^T \boldsymbol{f}} \quad (14)$$

Let $\tilde{\boldsymbol{f}}^T = \boldsymbol{f}^T(\boldsymbol{f}^T\boldsymbol{f})^{-1/2} = [\tilde{f}_1, \tilde{f}_2, \ldots, \tilde{f}_n]$, then we have $\tilde{\boldsymbol{f}}^T\tilde{\boldsymbol{f}} = 1$. Then, the Rayleigh quotient can be converted to $R(\mathbf{A}, \tilde{\boldsymbol{f}}) = \tilde{\boldsymbol{f}}^T\mathbf{A}\tilde{\boldsymbol{f}}$. Rayleigh-Ritz Theorem gives the minimum and maximum of Rayleigh quotient [20].

**Rayleigh-Ritz Theorem.** Given the sorted eigenvalues $\lambda_1 \leq \lambda_2 \leq \cdots \leq \lambda_n$ of $\mathbf{A}$, we have $\lambda_1 \leq R(\mathbf{A}, \tilde{\boldsymbol{f}}) \leq \lambda_n$.

**Proof.** Given the diagonal matrix $\mathbf{\Lambda} = \text{diag}(\lambda_1, \lambda_2, \ldots, \lambda_n)$ and the matrix $\mathbb{V} = [\mathbb{v}_1, \mathbb{v}_2, \ldots, \mathbb{v}_n]$ composed of the corresponding orthonormal eigenvectors, $\mathbf{A}$ can be decomposed as $\mathbf{A} = \mathbb{V}^T \mathbf{\Lambda} \mathbb{V}$, we have

$$R(\mathbf{A}, \tilde{\boldsymbol{f}}) = \tilde{\boldsymbol{f}}^T \mathbb{V}^T \mathbf{\Lambda} \mathbb{V} \tilde{\boldsymbol{f}} = \sum_{i=1}^{n} \lambda_i \tilde{f}_i^2 \mathbb{v}_i^T \mathbb{v}_i = \sum_{i=1}^{n} \lambda_i \tilde{f}_i^2 \quad (15)$$

$$\lambda_1 = \lambda_1 \sum_{i=1}^{n} \tilde{f}_i^2 \leq R(\mathbf{A}, \tilde{\boldsymbol{f}}) \leq \lambda_n \sum_{i=1}^{n} \tilde{f}_i^2 = \lambda_n \quad (16)$$

We can further obtain Theorem 1 when $\mathbf{A}$ is the Laplacian matrix and $\tilde{\boldsymbol{f}}$ is a unit indicator vector based on the Rayleigh-Ritz Theorem

**Theorem 1.** Given Laplacian $\mathbf{L}$, its eigenvalues $\lambda_1 \leq \lambda_2 \leq \cdots \leq \lambda_n$ and $k$ orthonormal indicator vectors $\tilde{\mathbf{x}}_1, \tilde{\mathbf{x}}_2, \ldots, \tilde{\mathbf{x}}_k$, we have $\sum_{i=1}^{k} R(\mathbf{L}, \tilde{\mathbf{x}}_i) \geq \sum_{i=1}^{k} \lambda_i$.

**Proof.** We can use Eq. (15) to convert the Rayleigh quotient to

$$\sum_{i=1}^{k} R(\mathbf{L}, \tilde{\mathbf{x}}_i) = \sum_{i=1}^{k} \tilde{\mathbf{x}}_i^T \mathbf{L} \tilde{\mathbf{x}}_i = \sum_{i=1}^{k} \sum_{j=1}^{n} \lambda_j \tilde{x}_{ij}^2 \quad (17)$$

In the normalized indicator vectors, the element values are transformed from 0 and 1 in Eq. (6) to 0 and $\frac{1}{\sqrt{|\mathbf{C}_i|}}$. Thus, we have

$$\sum_{i=1}^{k} \sum_{j=1}^{n} \lambda_j x_{ij}^2 = \sum_{i=1}^{k} \frac{\sum_{v_j \in \mathbf{C}_i} \lambda_j}{|\mathbf{C}_i|} \geq \sum_{i=1}^{k} \min_{v_j \in \mathbf{C}_i} \lambda_j \geq \sum_{i=1}^{k} \lambda_i \quad (18)$$

Theorem 1 indicates that the solution of Rcut problem approximately equals to the sum of the $k$ smallest eigenvalues of $\mathbf{L}$. Ncut can also be solved using the Rayleigh quotient. Firstly, we generalize the Rayleigh quotient to

$$R(\mathbf{A}, \mathbf{B}, \boldsymbol{f}) = \frac{\boldsymbol{f}^T \mathbf{A} \boldsymbol{f}}{\boldsymbol{f}^T \mathbf{B} \boldsymbol{f}} \quad (19)$$

where $\mathbf{B} \in \mathbb{R}^{n \times n}$ is a real symmetric matrix. Let $\boldsymbol{h} = \mathbf{B}^{1/2}\boldsymbol{f}$, then we have

$$R(\mathbf{A}, \mathbf{B}, \boldsymbol{f}) = R(\mathbf{B}^{-1/2}\mathbf{A}\mathbf{B}^{-1/2}, \boldsymbol{h}) = \frac{\boldsymbol{h}^T(\mathbf{B}^{-1/2}\mathbf{A}\mathbf{B}^{-1/2})\boldsymbol{h}}{\boldsymbol{h}^T\boldsymbol{h}} \quad (20)$$

As described by Rayleigh quotient theorems, the Ncut problem is approximately equivalent to calculating the sum of the $k$ smallest eigenvalues of $\mathbf{D}^{-1/2}\mathbf{L}\mathbf{D}^{-1/2}$. The initial indicator $\mathbf{X}$ is accordingly solved as a matrix composed of the leading $k$ eigenvectors. Consequently, $\mathbf{X}$ loses the binary indicative information, and postprocessing is needed to obtain a discrete clustering. In addition, Laplacian decomposition has high complexity in time and space [13-14].

## 3 FUNDAMENTAL ALGORITHM

In this section, we propose a novel graph cut function MeanCut, and conduct a greedy optimization. Since it utilizes pointwise agglomerative clustering rather than spectral decomposition, the indicator vectors can be solved nondestructively.

### 3.1 Path-based Similarity

Traditional spectral clustering algorithms commonly construct the similarity matrix using Euclidean distance measurement. However, this way cannot guarantee the strong associations between the points in the same cluster (shown in Fig. 1(a)). We hence adopt path-based similarity to enhance intra-cluster associations and make the inter-cluster separation more distinct.

In a fully connected graph, many communication paths exist between two points. We denote all the paths between $v_i$ and $v_j$ as $\mathbf{P}_{ij}$. Path $p$ contains multiple vertices that can be arranged from the start to the end vertex as $v_{\langle 1 \rangle}, v_{\langle 2 \rangle}, \ldots, v_{\langle m \rangle}$. The minimum edge weight of a path is treated as the path-based weight of $v_i$ and $v_j$ in the path. We define the path having the maximum path-based weight as the optimal path (OP) between $v_i$ and $v_j$, and the corresponding weight as their path-based similarity

$$\widehat{w}_{ij} = \max_{p \in \mathbf{P}_{ij}} \min_{v_{\langle s \rangle} \in p} w_{\langle s \rangle \langle s+1 \rangle} \quad (21)$$

To measure the path-based similarities, we utilize the Floyd-Warshall algorithm to seek for the OPs of all point pairs and the workflow is illustrated in Fig. 1(b). In a fully connected graph, the OPs are initialized as the edges that connect two points directly. The path-based weights and generated OPs will be updated in each iteration. We select $v_t$ as the transition vertex in the $t$th iteration. For two vertices $v_r$ and $v_s$, if $\widehat{w}_{rs}^{t-1} < \min(\widehat{w}_{rt}^{t-1}, \widehat{w}_{ts}^{t-1})$, their path-based weight and the path $op_{rs}^{t-1}$ will be updated as

$$\widehat{w}_{rs}^{t} = \min(\widehat{w}_{rt}^{t-1}, \widehat{w}_{ts}^{t-1}) \tag{22}$$

$$OP_{rs}^{t} = OP_{rt}^{t-1} \oplus OP_{ts}^{t-1} \tag{23}$$

where $\oplus$ denotes the operation to merge two paths.

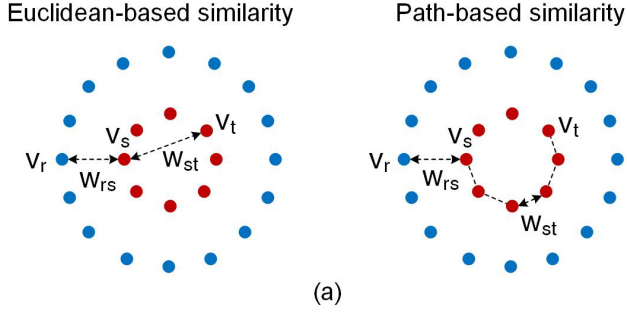

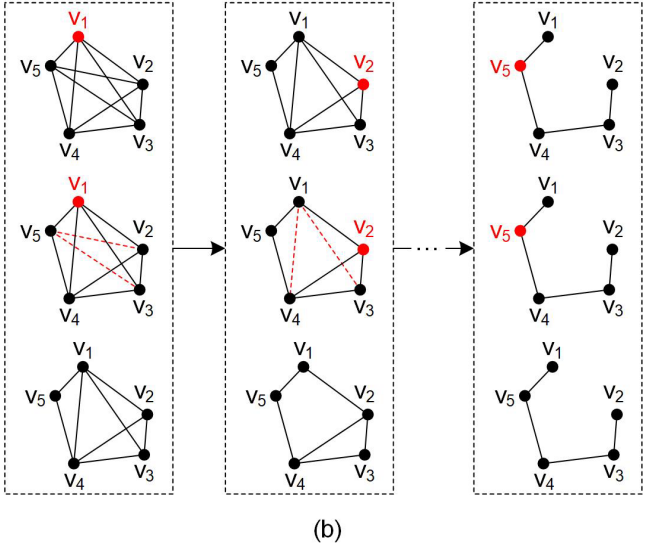

Fig. 1. Illustration of the path-based similarity. (a) An example shows two different clusters with red and blue points respectively. $v_r$ belongs to the blue cluster, while $v_s$ and $v_t$ belong to the red cluster. For the Euclidean-based similarity, $w_{rs} > w_{st}$, but for the path-based similarity, $w_{rs} < w_{st}$. (b) Workflow of the Floyd-Warshall algorithm to search for OPs, where the red point denotes the transition vertex and the red dashed lines represent the useless paths that are deleted in each iteration.

### 3.2 MeanCut Function

Indicator matrix represents the subordinate relations between the points and clusters. It is introduced to assist the definition of graph cut in spectral clustering. However, classical spectral clustering algorithms relax the elements in the indicator vectors from a discrete binary to a continuous space. Consequently, the solved indicator matrix loses the ability to map the points to clusters directly. Rcut and Ncut adopt K-means to restore the information, but errors and bias are also introduced.

In this paper, we intend to solve the indicator matrix using pointwise optimization rather than global relaxation. However, the graph cuts, e.g., Rcut and Ncut, are not suitable to be an objective function that can be solved in a greedy manner, since they are global-optimization-oriented. Hence, we define a new graph cut function Mean-Cut as

$$MeanCut(\mathbf{x}_i) = \frac{n}{n - \mathbf{x}_i^T \mathbf{x}_i} \cdot \frac{\mathbf{x}_i^T \mathbf{L} \mathbf{x}_i}{\mathbf{x}_i^T \mathbf{D} \mathbf{x}_i} \tag{24}$$

It is improved based on Ncut by standardizing the weight sum of edges between different clusters and degree sum of all points respectively. This modification facilitates to evaluate whether the MeanCut function decreases when adding one point into the cluster in a greedy way. Specifically, we traverse all unlabeled points and add them to the cluster point-by-point iteratively. In the $i$th iteration, vertex $v_j$ is added to the current cluster $\mathbf{C}_i$, and $x_{ij}$ in the indicator vector $\mathbf{x}_i$ will be assigned to 1. If this assignment makes MeanCut smaller, the corresponding indicator vector $\mathbf{x}_i$ will accept the above update. One cluster is generated after each iteration, and the iterations end until all points are labeled.

### 3.3 Degree Descent Optimization

The order to traverse the points matters, since different input orders produce different trends of MeanCut and clustering results.

Degree descent is the best order to optimize MeanCut greedily when using the path-based similarity. To prove this proposition, we put forward three assumptions

**Assumption 1.** For any two different clusters $\mathbf{C}_i$ and $\mathbf{C}_j$, if points $v_r, v_s \in \mathbf{C}_i$ and $v_t \in \mathbf{C}_j$, then

$$\widehat{w}_{rs} \geq \max(\widehat{w}_{rt}, \widehat{w}_{st}) \tag{25}$$

**Assumption 2.** If points $v_r, v_s \in \mathbf{C}_i$, then

$$\widehat{w}_{rs} \geq \max\left(\frac{d_r}{n}, \frac{d_s}{n}\right) \tag{26}$$

**Assumption 3.** If points $v_r, v_s, v_t$ belong to cluster $\mathbf{C}_i$, then

$$\widehat{w}_{rs} - \frac{d_r}{n} \geq \widehat{w}_{rt} - \widehat{w}_{st} \tag{27}$$

The raised three assumptions are reasonable in most cases under the measurement of path-based similarity. Assumption 1 means that a point has higher path-based similarity with points in the same cluster than points in different clusters. Assumption 2 considers that the similarity between a point and any other point in the same cluster is greater than its average similarity with all points. Assumption 3 considers the difference between two intra-cluster similarities are smaller than that between similarity of any point pair in the same cluster and the average similarity with all points.

Based on the three assumptions, we can obtain two lemmas:

**Lemma 1.** Suppose the points $v_1, v_2, \ldots, v_m$ are in the same cluster $\mathbf{C}_i$, and $\mathbf{x}_i$ is the corresponding indicator vector, we

have

$$\frac{\mathbf{x}_i^T \mathbf{W} \mathbf{x}_i}{\mathbf{x}_i^T \mathbf{L} \mathbf{x}_i} \geq \frac{m}{n-m} \quad (34)$$

**Proof.** According to Assumption 2, we can obtain

$$\sum_{i=1}^m \widehat{w}_{ij} \geq \frac{m}{n} d_i \Rightarrow \sum_{i=1}^m \sum_{j=1}^m \widehat{w}_{ij} \geq \frac{m}{n} \sum_{i=1}^m d_i$$

$$\Rightarrow \mathbf{x}_i^T \mathbf{W} \mathbf{x}_i \geq \frac{m}{n} \mathbf{x}_i^T \mathbf{D} \mathbf{x}_i$$

$$\Rightarrow \mathbf{x}_i^T \mathbf{W} \mathbf{x}_i \geq \frac{m}{n} (\mathbf{x}_i^T \mathbf{W} \mathbf{x}_i + \mathbf{x}_i^T \mathbf{L} \mathbf{x}_i)$$

$$\Rightarrow \frac{\mathbf{x}_i^T \mathbf{W} \mathbf{x}_i}{\mathbf{x}_i^T \mathbf{L} \mathbf{x}_i} \geq \frac{m}{n-m} \quad (35)$$

**Lemma 2.** Suppose points $v_1, v_2, \ldots, v_m$ are in the same cluster $\mathbf{C}_i$, and $\mathbf{x}_i$ is the corresponding indicator vector, we also have

$$2mn \sum_{i=1}^m \widehat{w}_{i\,m+1} \geq (n+m) \mathbf{x}_i^T \mathbf{W} \mathbf{x}_i + m \mathbf{x}_i^T \mathbf{L} \mathbf{x}_i \quad (36)$$

**Proof.** According to Assumption 3, we can obtain

$$\widehat{w}_{i\,m+1} - \frac{1}{n} d_i \geq \frac{1}{m} \sum_{j=1}^m \widehat{w}_{ij} - \widehat{w}_{i\,m+1}$$

$$\Rightarrow \sum_{i=1}^m \widehat{w}_{i\,m+1} - \frac{1}{n} \mathbf{x}_i^T \mathbf{D} \mathbf{x}_i \geq \frac{1}{m} \mathbf{x}_i^T \mathbf{W} \mathbf{x}_i - \sum_{i=1}^m \widehat{w}_{i\,m+1}$$

$$\Rightarrow 2mn \sum_{i=1}^m \widehat{w}_{i\,m+1} \geq (n+m) \cdot \mathbf{x}_i^T \mathbf{W} \mathbf{x}_i + m \mathbf{x}_i^T \mathbf{L} \mathbf{x}_i \quad (37)$$

Using the above two lemmas, we can generalize a theorem

**Degree Descent Theorem.** Suppose $v_1, v_2, \ldots, v_m$ are in the same cluster $\mathbf{C}_i$, their degrees satisfy $d_1 \geq d_2 \geq \cdots \geq d_m$ and the generated indicator vector is $\mathbf{x}_i^m$. If $v_{m+1}$ also belongs to $\mathbf{C}_i$ and its degree $d_{m+1} \leq d_m$, adding $v_{m+1}$ to $\mathbf{C}_i$ will transform the indicator vector to $\mathbf{x}_i^{m+1}$ and decrease the MeanCut

$$MeanCut(\mathbf{x}_i^{m+1}) \leq MeanCut(\mathbf{x}_i^m) \quad (38)$$

**Proof.** The original proposition is equivalent to

$$\frac{n}{n-(m+1)} \cdot \frac{(\mathbf{x}_i^{m+1})^T \mathbf{L} \mathbf{x}_i^{m+1}}{(\mathbf{x}_i^{m+1})^T \mathbf{D} \mathbf{x}_i^{m+1}} \leq \frac{n}{n-m} \cdot \frac{(\mathbf{x}_i^m)^T \mathbf{L} \mathbf{x}_i^m}{(\mathbf{x}_i^m)^T \mathbf{D} \mathbf{x}_i^m}$$

$$\Leftrightarrow (n-m)(\mathbf{x}_i^m)^T \mathbf{D} \mathbf{x}_i^m \cdot (\mathbf{x}_i^{m+1})^T (\mathbf{D} - \mathbf{W}) \mathbf{x}_i^{m+1}$$
$$\leq (n-m-1) \cdot (\mathbf{x}_i^{m+1})^T \mathbf{D} \mathbf{x}_i^{m+1} \cdot (\mathbf{x}_i^m)^T (\mathbf{D} - \mathbf{W}) \mathbf{x}_i^m$$

$$\Leftrightarrow (\mathbf{x}_i^{m+1})^T \mathbf{D} \mathbf{x}_i^{m+1} \cdot [(\mathbf{x}_i^m)^T \mathbf{D} \mathbf{x}_i^m + (n-m-1) \cdot (\mathbf{x}_i^m)^T \mathbf{W} \mathbf{x}_i^m]$$
$$\leq (n-m) \cdot (\mathbf{x}_i^m)^T \mathbf{D} \mathbf{x}_i^m \cdot (\mathbf{x}_i^{m+1})^T \mathbf{W} \mathbf{x}_i^{m+1}$$

$$\Leftrightarrow \frac{(\mathbf{x}_i^{m+1})^T \mathbf{D} \mathbf{x}_i^{m+1}}{(\mathbf{x}_i^m)^T \mathbf{D} \mathbf{x}_i^m} \leq \frac{(n-m) \cdot (\mathbf{x}_i^{m+1})^T \mathbf{W} \mathbf{x}_i^{m+1}}{(\mathbf{x}_i^m)^T \mathbf{D} \mathbf{x}_i^m + (n-m-1) \cdot (\mathbf{x}_i^m)^T \mathbf{W} \mathbf{x}_i^m} \quad (39)$$

According to $d_1 \geq d_2 \geq \cdots \geq d_m \geq d_{m+1}$, we have

$$\frac{(\mathbf{x}_i^{m+1})^T \mathbf{D} \mathbf{x}_i^{m+1}}{m+1} \leq \frac{(\mathbf{x}_i^m)^T \mathbf{D} \mathbf{x}_i^m}{m} \Rightarrow \frac{(\mathbf{x}_i^{m+1})^T \mathbf{D} \mathbf{x}_i^{m+1}}{(\mathbf{x}_i^m)^T \mathbf{D} \mathbf{x}_i^m} \leq \frac{m+1}{m} \quad (40)$$

Therefore, the proof of inequality (39) can be converted to

$$\frac{m+1}{m} \leq \frac{(n-m) \cdot (\mathbf{x}_i^{m+1})^T \mathbf{W} \mathbf{x}_i^{m+1}}{(\mathbf{x}_i^m)^T \mathbf{D} \mathbf{x}_i^m + (n-m-1) \cdot (\mathbf{x}_i^m)^T \mathbf{W} \mathbf{x}_i^m}$$

$$\Leftrightarrow (m+1) \cdot [(\mathbf{x}_i^m)^T (\mathbf{L} + \mathbf{W}) \mathbf{x}_i^m + (n-m-1) \cdot (\mathbf{x}_i^m)^T \mathbf{W} \mathbf{x}_i^m]$$
$$\leq m(n-m) \cdot ((\mathbf{x}_i^m)^T \mathbf{W} \mathbf{x}_i^m + 2 \sum_{i=1}^m \widehat{w}_{i\,m+1} + 1)$$

$$\Leftrightarrow (m+1) \cdot (\mathbf{x}_i^m)^T \mathbf{L} \mathbf{x}_i^m + (n-m) \cdot (\mathbf{x}_i^m)^T \mathbf{W} \mathbf{x}_i^m$$
$$\leq 2m(n-m) \cdot \sum_{i=1}^m \widehat{w}_{i\,m+1} + m(n-m) \quad (41)$$

If $(\mathbf{x}_i^m)^T \mathbf{W} \mathbf{x}_i^m \leq m \sum_{i=1}^m \widehat{w}_{i\,m+1}$, according to Lemma 1, we have

$$(m+1) \cdot (\mathbf{x}_i^m)^T \mathbf{L} \mathbf{x}_i^m + (n-m) \cdot (\mathbf{x}_i^m)^T \mathbf{W} \mathbf{x}_i^m$$

$$\leq \frac{(m+1)(n-m)}{m} \cdot (\mathbf{x}_i^m)^T \mathbf{W} \mathbf{x}_i^m + (n-m) \cdot (\mathbf{x}_i^m)^T \mathbf{W} \mathbf{x}_i^m$$

$$\leq (2m+1)(n-m) \cdot \sum_{i=1}^m \widehat{w}_{i\,m+1}$$

$$\leq 2m(n-m) \cdot \sum_{i=1}^m \widehat{w}_{i\,m+1} + m(n-m) \quad (42)$$

Hence, inequality (41) holds. While in another case of $(\mathbf{x}_i^m)^T \mathbf{W} \mathbf{x}_i^m > m \sum_{i=1}^m \widehat{w}_{i\,m+1}$, based on Lemma 2, we have

$$(m+1) \cdot (\mathbf{x}_i^m)^T \mathbf{L} \mathbf{x}_i^m + (n-m) \cdot (\mathbf{x}_i^m)^T \mathbf{W} \mathbf{x}_i^m$$

$$= \frac{m+1}{m} \cdot [m(\mathbf{x}_i^m)^T \mathbf{L} \mathbf{x}_i^m + (n+m) \cdot (\mathbf{x}_i^m)^T \mathbf{W} \mathbf{x}_i^m]$$

$$- \frac{2m^2 + m + n}{m} \cdot (\mathbf{x}_i^m)^T \mathbf{W} \mathbf{x}_i^m$$

$$\leq 2n(m+1) \cdot \sum_{i=1}^m \widehat{w}_{i\,m+1} - \frac{2m^2 + m + n}{m} \cdot (\mathbf{x}_i^m)^T \mathbf{W} \mathbf{x}_i^m$$

$$< 2n(m+1) \cdot \sum_{i=1}^m \widehat{w}_{i\,m+1} - (2m^2 + m + n) \cdot \sum_{i=1}^m \widehat{w}_{i\,m+1}$$

$$= (2m+1)(n-m) \cdot \sum_{i=1}^m \widehat{w}_{i\,m+1}$$

$$\leq 2m(n-m) \cdot \sum_{i=1}^m \widehat{w}_{i\,m+1} + m(n-m) \quad (43)$$

According to this theorem, if the data distribution satisfies the three assumptions, MeanCut would decrease monotonically when we agglomerate the points belonging to the same cluster in the descending order of degree.

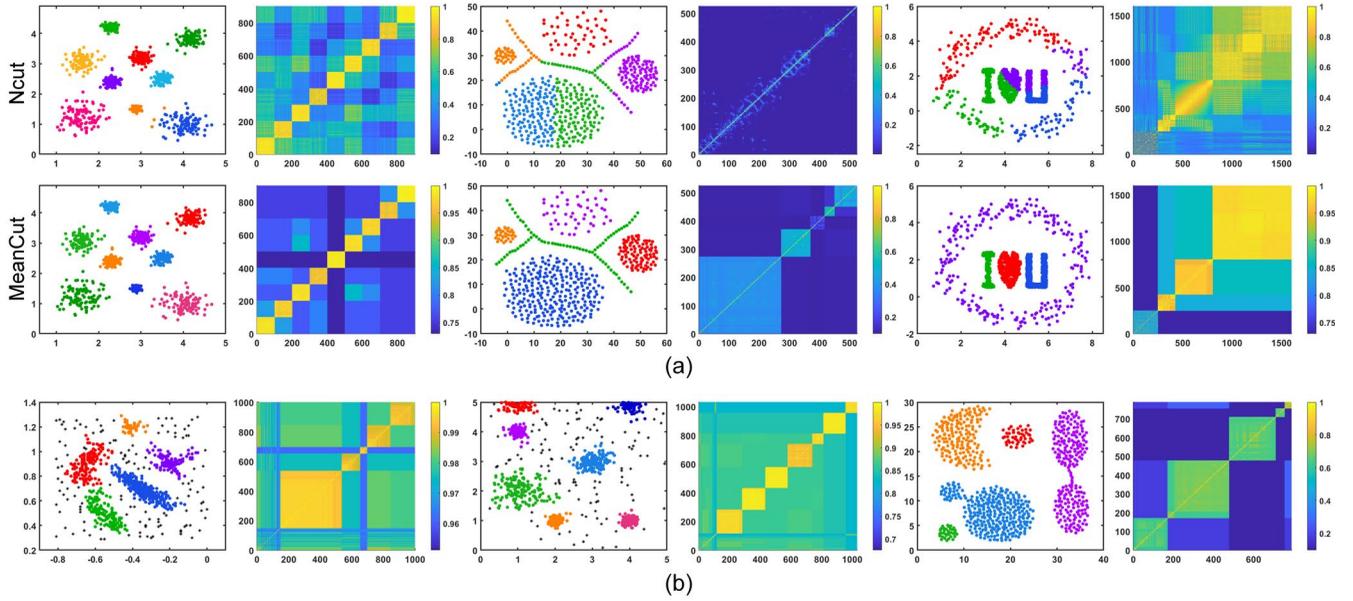

Fig. 2. Clustering results and similarity matrices of MeanCut on synthetic datasets. (a) Comparison with Ncut on DS1-DS3 datasets. (b) Performances of MeanCut on DS4 and DS5 datasets with noise (noise threshold is set as 20 and 50 respectively), and DS6 having weakly connected clusters.

### 3.4 Workflow of MeanCut Clustering

Based on the Degree Descent Theorem, a parameter-free MeanCut clustering algorithm is proposed by adopting a greedy strategy. The workflow is presented in Algorithm 1. Specifically, we first calculate the degree matrix **D** and Laplacian **L** according the path-based similarity matrix **W**, and arrange all points in the descending order of degree. We add the unassigned points to one cluster point-by-point from the sorted vertices **V**. MeanCut function is updated accordingly every time when a point is added. If MeanCut decreases, the point is formally assigned to the current cluster; otherwise, it is removed from the cluster, and move to next unassigned point. Each round of traversing generates one cluster, and it ends until all points are assigned.

### 3.5 Performance on Synthetic Datasets

To evaluate the effectiveness, we benchmark our algorithm on six synthetic datasets with different shaped clusters in Fig. 2. Ncut and MeanCut have similar performance on DS1 with nine spherical clusters, but Ncut misidentified three points of the cluster in the bottom right corner. DS2 contains four spherical clusters and a path-shaped cluster. Ncut cannot detect the non-spherical cluster using Euclidean-based similarity, and in turn interfere the identification of the spherical clusters. However, the path-based similarity matrix presents clearer block structures and benefits MeanCut to extract each cluster accurately. DS3 is more challenging in data distribution, since it contains three dense clusters surrounded by a ring-shaped cluster with a significant density difference. MeanCut outperforms Ncut on such an island-shaped distribution.

Meanwhile, MeanCut is robust to the noise as shown in the results of DS4 and DS5. MeanCut assigns all points

---

**Algorithm 1** Parameter-free MeanCut Clustering

*Input:* vertices **V** and path-based similarity matrix **W**
1: Calculate the degree matrix **D** and Laplacian **L**
2: Arrange **V** as $\{v_1, v_2, \ldots, v_n\}$ in degree descending order
3: **for** each cluster $\mathbf{C}_i$
4:    Initialize $\mathbf{x}_i^T = [x_{i1}, x_{i2}, \ldots, x_{in}] = [0,0,\ldots,0]$
5:    Add the first unassigned point $v_j$ to $\mathbf{C}_i$ ($x_{ij} = 1$)
6:    Initialize $MeanCut\,(\mathbf{x}_i) = \frac{n}{n - \mathbf{x}_i^T \mathbf{x}_i} \cdot \frac{\mathbf{x}_i^T \mathbf{L} \mathbf{x}_i}{\mathbf{x}_i^T \mathbf{D} \mathbf{x}_i}$
7:    **for** the next unassigned point $v_k$ in **V**
8:      Add $v_k$ to $\mathbf{C}_i$ ($x_{ik} = 1$)
9:      Calculate MeanCut as $MeanCut\,(\mathbf{x}_i)'$
10:     **if** $MeanCut\,(\mathbf{x}_i)' \leq MeanCut\,(\mathbf{x}_i)$
11:       Update MeanCut to $MeanCut\,(\mathbf{x}_i)'$
12:     **else**
13:       Remove $v_k$ from $\mathbf{C}_i$ ($x_{ik} = 0$)
14:     **end if**
15:    **end for**
16: **end for**
17: **return** $\mathbf{x}_1, \mathbf{x}_2, \ldots, \mathbf{x}_k$

---

to clusters including the noise. Due to the dispersed distribution, each noisy cluster only contains a small number of noise points. Hence, a noise threshold is qualified to eliminate the clusters composed of limited noise points. The optimal threshold can be specified in the range between the number of points for the smallest true cluster and the largest noisy cluster.

However, the result on DS6 indicates that MeanCut is unable to separate the weakly connected clusters. Because the points at the junction areas of the two clusters constitute communication paths between them. In this case, the weakly connected clusters have high path-based similarities, which make them hard to separate.

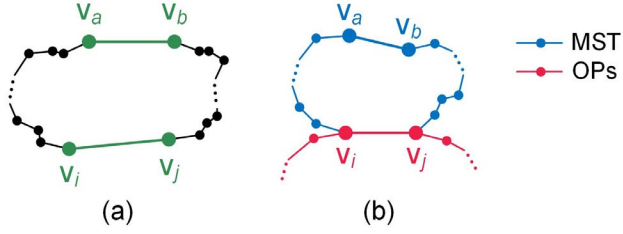

Fig. 3. Examples for proving that the union of all OPs is equivalent to the maximum spanning tree (MST). (a) A loop formed by OPs. (b) The union of all OPs (red) and the MST (blue).

## 4 IMPROVEMENT STRATEGIES

MeanCut relies on the calculation of path-based similarity by searching OPs as discussed in Section 3.1. However, finding the OPs of all point pairs using Floyd-Warshall algorithm has high time complexity of $O(n^3)$. Meanwhile, the path-based similarity enhances the associations between the weakly connected clusters that makes MeanCut difficult to separate them.

In this section, we prove that generating the maximum spanning tree (MST) is equivalent to the search of OPs, and propose a FastMST algorithm based on DBSCAN [21] and Kruskal algorithms [22] that can reduce the time complexity to $O(n\log n)$ in the best case. Besides, considering that the junction points between clusters are in local density saddles, a density gradient factor is defined to detect them.

### 4.1 Similarity Calculation using MST

The OP of two points refers to the one whose minimum edge weight is the maximum among all paths between the two points. MST is the spanning tree whose sum of all edge weights is maximum among all spanning trees. We can conclude the following theorem

**Theorem 2.** Given the union of OPs of all point pairs $\bigcup_{i,j=1}^{n} \text{OP}_{ij}$ and the MST(**G**) of graph **G**, and suppose there are no equivalent edges (refers to the edges having the same minimum weights in a loop of the graph, and deleting any of them would not affect the path-based similarity of any point pair), we have

$$\bigcup_{i,j=1}^{n} \text{OP}_{ij} = \text{MST}(\mathbf{G}) \qquad (44)$$

**Proof.** We first prove that $\bigcup_{i,j=1}^{n} \text{OP}_{ij}$ is a spanning tree, i.e., it can connect any two points and has no loops without considering the equivalent edges. Using the method of reductio ad absurdum, we assume that there is a loop composed of multiple OPs in Fig. 3(a). This loop goes through two vertices $v_i$ and $v_j$, and we have $\hat{w}_{ij} = w_{ij}$ according to Floyd-Warshall algorithm. We assume the minimum edge weight of another path $p_{i \to a \to b \to j}$ is $\hat{w}_{ab}$, so $p_{ab}$ must be the OP of $v_a$ and $v_b$. If $\hat{w}_{ij} = \hat{w}_{ab}$, $p_{ij}$ and $p_{ab}$ are two equivalent edges that do not need to be considered. If $\hat{w}_{ij} > \hat{w}_{ab}$, $p_{ab}$ can be replaced with $p_{a \to i \to j \to b}$ that makes the path-based similarity of $v_a$ and $v_b$ larger, which conflicts with that $p_{ab}$ is the OP of $v_a$ and $v_b$. If $\hat{w}_{ij} < \hat{w}_{ab}$, $p_{ij}$ can be replaced by $p_{i \to a \to b \to j}$, which also leads to conflicts. Hence, $\bigcup_{i,j=1}^{n} \text{OP}_{ij}$ is a spanning tree.

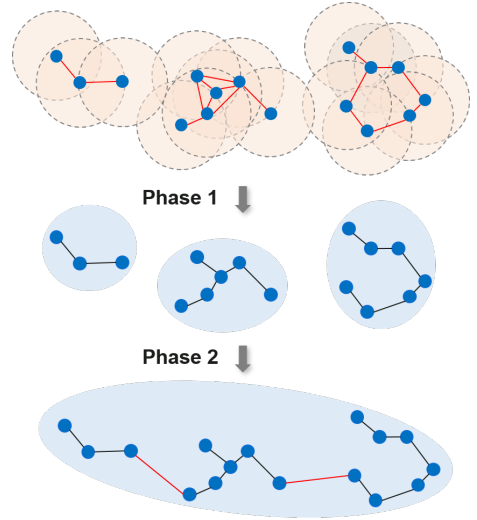

Fig. 4. Workflow of the FastMST algorithm.

Then, we intend to prove that $\bigcup_{i,j=1}^{n} \text{OP}_{ij}$ is a MST. We also utilize the method of reductio ad absurdum, and assume that $\bigcup_{i,j=1}^{n} \text{OP}_{ij}$ is not the MST. In this case, there are edges in $\bigcup_{i,j=1}^{n} \text{OP}_{ij}$ and MST do not coincide. As shown in Fig. 3(b), the vertices $v_i$ and $v_j$ is connected by two paths, one is their OP, and the other is a reachable subtree of MST. We assume $p_{ab}$ is the edge with the minimum weight in the subtree. If $\hat{w}_{ij} = \hat{w}_{ab}$, $p_{ij}$ and $p_{ab}$ are two equivalent edges that do not need to be considered. If $\hat{w}_{ij} > \hat{w}_{ab}$, the weight sum of all edges in the MST would become larger by replacing $p_{ab}$ with $p_{ij}$, which causes a contradiction with the maximality of MST. While, if $\hat{w}_{ij} < \hat{w}_{ab}$, $p_{ij}$ can be substituted by $p_{i \to a \to b \to j}$, which conflicts with that $p_{ab}$ is the OP between $v_i$ and $v_j$. Therefore, the union of all OPs is the MST.

### 4.2 FastMST Algorithm

Based on Theorem 2, we can calculate the path-based similarity by generating the MST rather than finding the OPs. Nonetheless, it is still insufficient to handle large dataset in an efficient manner using the classical MST algorithms, e.g., Boruvka [23], Kruskal and Prim algorithms [24]. Because they have time complexity of $O(n^2 \log n)$ in a fully connected graph. Thus, we propose FastMST algorithm to further accelerate the computation.

The workflow of FastMST algorithm is illustrated in Fig. 4. Initially, we adopt DBSCAN to extract connected subgraphs and remove the edges with length larger than the distance radius $Eps$. Then, a two-phase Kruskal algorithms would be performed. In the first phase, it generates a subMST for each subgraph using the remaining edges. In the second phase, we take each connected subgraph as a new vertex, and apply Kruskal algorithm to all cross-sub-

**Algorithm 2** FastMST Algorithm

*Input:* the vertices **V** and $ratio$ of $Eps$
1: Calculate the diagonal length $\ell$ of MBR of **V**
2: Calculate $Eps = \ell \cdot ratio$
4: Run DBSCAN on vertices **V** with $Eps$ to extract subgraphs $\mathbf{G}_1, \mathbf{G}_2, \dots, \mathbf{G}_m$
5: **for** each subgraph $\mathbf{G}_i$
6:     Run Kruskal's algorithm to generate MST($\mathbf{G}_i$)
7: **end**
8: Run Kruskal's algorithm on the removed cross-subgraph edges
9: **return** the global MST(**G**);

graph edges that have been removed by DBSCAN. Specifically, we arrange these edges in descending order of edge weight, and add them to the global MST one by one to connect all points without creating a loop. The pseudocode of FastMST is presented in Algorithm 2. In practice, considering $Eps$ varies with data distributions, we adopt a percentile $ratio$ to determine $Eps$ as the diagonal length of the maximum bounding rectangle (MBR) times $ratio$.

### 4.3 Complexity of FastMST Algorithm

To assess the computational efficiency of FastMST, the time complexity is analyzed. The time consumption of FastMST is composed of two parts, including DBSCAN and twice Kruskal algorithms. DBSCAN has time complexity of $O(n\log n)$, and that of Kruskal algorithm is $O(|\mathbf{E}|\log|\mathbf{E}|)$ relating to the number of edges. We assume that $m$ subgraphs have been extracted by DBSCAN, and they contain $a_1, a_2, \dots, a_m$ vertices respectively. Here, we suppose $a_1 \leq a_2 \dots \leq a_m$. We set DBSCAN not to detect any noise (i.e., $minPts$=1), and the number of vertices satisfies $\sum_{i=1}^m a_i = n$. Thus, the complexity of twice Kruskal algorithms in two phases can be obtained:

**Phase 1.** For subgraph $\mathbf{G}_i$, the number of remaining edges is $a_i - 1$ in the best case, and $\frac{a_i(a_i-1)}{2}$ in the worst case. Thus, the best complexity of phase 1 can be denoted as

$$O(\mathrm{T}_1^{best}) = O(\sum_{i=1}^m a_i \log a_i) = O(\log \prod_{i=1}^m a_i^{a_i})$$

$$\Rightarrow O(\log a_1^{\sum_{i=1}^m a_i}) \leq O(\mathrm{T}_1^{best}) \leq O(\log a_m^{\sum_{i=1}^m a_i})$$

$$\Rightarrow O(n \log a_1) \leq O(\mathrm{T}_1^{best}) \leq O(n \log a_m) \quad (45)$$

Similarly, we can obtain the worst complexity

$$O(\mathrm{T}_1^{worst}) = O(\sum_{i=1}^m a_i^2 \log a_i) = O(\log \prod_{i=1}^m a_i^{a_i^2})$$

$$\Rightarrow O(\log a_1^{\sum_{i=1}^m a_i^2}) \leq O(\mathrm{T}_1^{worst}) \leq O(\log a_m^{\sum_{i=1}^m a_i^2}) \quad (46)$$

We suppose that each subgraph has the same number of vertices, then the complexity of $O(\mathrm{T}_1^{best})$ equals to $O(n \log \frac{n}{m})$ and that of $O(\mathrm{T}_1^{worst})$ is $O(\frac{n^2}{m} \log \frac{n}{m})$.

**Phase 2.** The number of removed cross-subgraph edges is

**Algorithm 3** Improved MeanCut Clustering

*Input:* the vertices **V**, $ratio$ of $Eps$, $K$ of KNN and a $percentile$ threshold of DGF
1: Search the KNN of **V**
2: Calculate the DGF and sort them in ascending order
3: **for** each point $v_i$
4:     **if** $\mathrm{DGF}(v_i) < \mathrm{DGF}_{percentile}$
5:        Add $v_i$ to ***junV***
6:     **else**
7:        Add $v_i$ to ***intV***
8:     **end**
9: **end**
10: Run FastMST on ***intV*** and calculate path-based **W**
11: Run parameter-free MeanCut clustering
12: Assign the label of the nearest ***intV*** to ***junV***
13: **return** the clustering labels

$$\frac{1}{2}\sum_{i=1}^m a_i(n-a_i) = \frac{1}{2}(n^2 - \sum_{i=1}^m a_i^2) \quad (47)$$

So, the complexity of phase 2 can be formulated

$$O(\mathrm{T}_2) = O((n^2 - \sum_{i=1}^m a_i^2)\log(n^2 - \sum_{i=1}^m a_i^2)) \quad (48)$$

The average $O(\mathrm{T}_2)$ is $O((1-\frac{1}{m})n^2 \log((1-\frac{1}{m})n^2))$. Thus, we can obtain the average overall complexity of the twice Kruskal algorithms in the best and worst cases

$$O(\mathrm{T}^{best}) = O(n\log\frac{n}{m} + (1-\frac{1}{m})n^2\log((1-\frac{1}{m})n^2)) \quad (49)$$

$$O(\mathrm{T}^{worst}) = O(\frac{n^2}{m}\log\frac{n}{m} + (1-\frac{1}{m})n^2\log((1-\frac{1}{m})n^2)) \quad (50)$$

If $m = 1$, the best overall complexity is $O(n\log n)$ and the worst is $O(n^2 \log n)$, since all vertices have been assigned to a subgraph by DBSCAN and there are no cross-subgraph edges. While $m > 1$, let $s = \frac{n}{m} \in [1, n)$, then we have

$$O(\mathrm{T}^{best}(s)) = O(n\log s + (n^2 - ns)\log(n^2 - ns)) \quad (51)$$

$$O(\mathrm{T}^{worst}(s)) = O(ns\log s + (n^2 - ns)\log(n^2 - ns)) \quad (52)$$

In the domain of definition, both of $O(\mathrm{T}^{best}(s))$ and $O(\mathrm{T}^{worst}(s))$ are decreasing monotonically, because their derivatives are negative. Hence, we have

$$O(\mathrm{T}^{best}(n)) < O(\mathrm{T}^{best}(s)) \leq O(\mathrm{T}^{worst}(s)) \leq O(\mathrm{T}^{best}(1))$$

$$\Rightarrow O(n\log n) < O(\mathrm{T}^{best}(s)) \leq O(\mathrm{T}^{worst}(s))$$

$$\leq O((n^2 - n)\log(n^2 - n)) \leq O(n^2 \log n^2) = O(2n^2 \log n) \quad (53)$$

According to above analysis, the overall complexity of FastMST is between $O(n\log n)$ and $O(n^2\log n)$. The highest efficiency is obtained when all vertices are just grouped into one subgraph by DBSCAN, since the number of remaining edges is minimal. While, the lowest efficiency $O(n^2 \log n)$ occurs and all edges need to be sorted when $ratio$ is close to 0 or 1, which approaches to performing Kruskal algorithm in the global fully-connected graph.

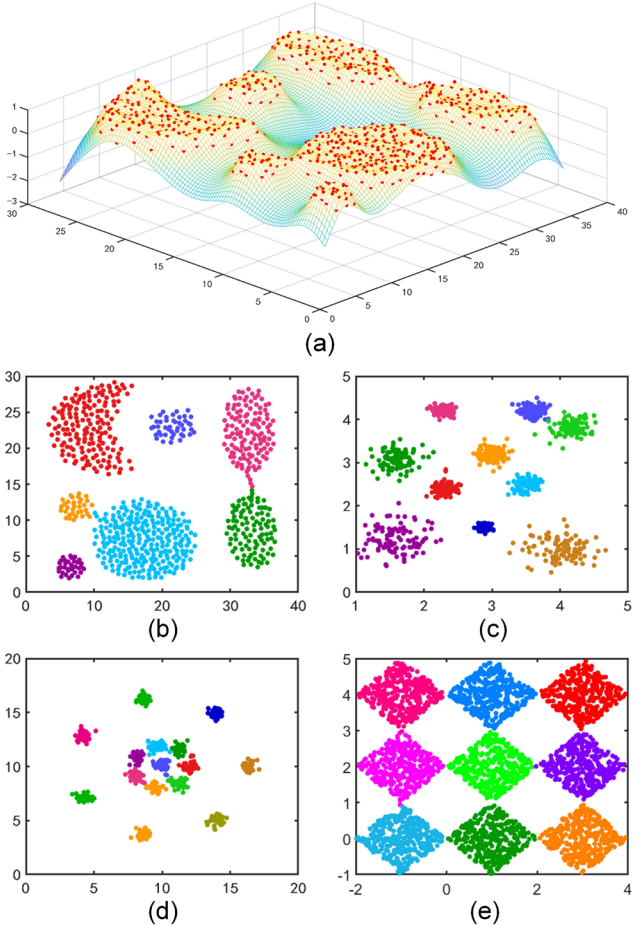

Fig. 5. Performances of the improved MeanCut on four synthetic datasets. (a) A 3D plot of DGF on DS6. (b)-(e) Clustering results on DS6-DS9.

### 4.4 Density Gradient Factor

To cope with weak connectivity, we design a factor to detect and eliminate the points at junction areas between weakly connected clusters before MeanCut clustering.

The junction points commonly have lower density than nearby intra-cluster points. That means they locate in local density saddles. We first define the density of point $v_i$ as the average distance from its KNN

$$density(v_i) = \frac{1}{K}\sum_{j=1}^{K}\|v_i - v_j\| \quad (54)$$

where $v_j$ is a membership of KNN of $v_i$.

Then, we define a density gradient factor (DGF) as the average density gradient

$$DGF(v_i) = \frac{1}{K}\sum_{j=1}^{K}\frac{density(v_j) - density(v_i)}{\|v_i - v_j\|} \quad (55)$$

In order to make sure $\|v_i - v_j\|$ does not equal to 0, the duplicate points are temporarily eliminated in the preprocessing step. We sort all DGF in an ascending order, and identify junction points by specifying a $percentile$ threshold of DGF.

TABLE 2
Information of 15 Real-world Datasets.

| Dataset | # of samples | Features | Classes |
|---|---|---|---|
| Iris | 150 | 4 | 3 |
| Wine | 178 | 13 | 3 |
| Seeds | 210 | 7 | 3 |
| Dermatology | 358 | 34 | 6 |
| Control | 600 | 60 | 6 |
| Breast-Cancer | 683 | 10 | 2 |
| Mice | 1,077 | 77 | 8 |
| COIL20 | 1,440 | 32×32 | 20 |
| Mfeat | 2,000 | 649 | 10 |
| Segment | 2,310 | 19 | 7 |
| Digits | 5,620 | 8×8 | 10 |
| Satimage | 6,430 | 36 | 7 |
| MNIST10k | 10,000 | 28×28 | 10 |
| PenDigits | 10,992 | 16 | 10 |
| Shuttle | 14,500 | 9 | 7 |

### 4.5 Workflow of Improved MeanCut

The workflow of the improved MeanCut is summarized in Algorithm 3. First, we calculate the DGF of all points, and divide the vertices **V** into internal and junction points, i.e., $int$**V** and $jun$**V** based on the sorted DGF. Then, for the $int$**V** points, we use FastMST to construct the path-based similarity matrix, and adopt the parameter-free MeanCut to generate initial clusters. The final cluster labels are obtained by assigning the labels of the nearest $int$**V** points to $jun$**V** points. In terms of the three parameters, $ratio$ only affects the time efficiency of FastMST, while the $k$ of KNN and $percentile$ threshold of DGF could influence the clustering accuracy.

We testify the improved MeanCut on four synthetic datasets (DS6-DS9) with weakly connected clusters. The 3D plot of DGF in Fig. 5a indicates that points at junction areas locate in local density saddles. Based on this characteristic, the improved algorithm using DGF identifies all the clusters accurately as shown in Fig. 5b-e, which demonstrate its effectiveness to handle weak connectivity.

## 5 EXPERIMENTS

In this section, we validate the clustering accuracy and time efficiency of our method on real-world and synthetic benchmarks. We also apply MeanCut on face recognition to evaluate its applicability and generality. All the experiments were conducted on a commodity desktop computer with an 8-core Intel i7 processor and 64 GB RAM.

### 5.1 Clustering Accuracy on Real-world Datasets

For accuracy evaluation, we benchmark MeanCut by comparing with six state-of-the-art clustering baselines on 15 real-world datasets.

TABLE 3
The Average ACC (±Standard Deviation) of Seven Clustering Algorithms on 15 Real-world Datasets
(The Best Result of Each Dataset is Highlighted in Bold).

| Dataset | K-means | Ncut | FastESC | SAMSC | U-SPEC | GCSED | MeanCut |
|---|---|---|---|---|---|---|---|
| Iris | 0.6067(±0.181) | 0.6133(±0.177) | 0.6457(±0.163) | 0.6314(±0.179) | 0.5695(±0.164) | 0.4790(±0.101) | **0.9070(±0.008)** |
| Wine | 0.6958(±0.182) | 0.6477(±0.194) | 0.6637(±0.186) | 0.6629(±0.179) | 0.6998(±0.180) | 0.6942(±0.181) | **0.9567(±0.004)** |
| Seeds | 0.6340(±0.173) | 0.5959(±0.189) | 0.5857(±0.199) | 0.5939(±0.192) | 0.6374(±0.177) | 0.6456(±0.169) | **0.9205(±0.002)** |
| Dermatology | 0.7039(±0.120) | 0.7709(±0.112) | 0.7019(±0.081) | 0.7945(±0.090) | 0.7298(±0.055) | 0.8212(±0.082) | **0.8580(±0.001)** |
| Control | 0.6293(±0.078) | 0.5793(±0.066) | 0.5288(±0.058) | 0.6381(±0.080) | **0.6857(±0.099)** | 0.5926(±0.057) | 0.6665(±0.007) |
| Breast-Cancer | 0.7356(±0.147) | 0.5984(±0.180) | 0.7971(±0.099) | 0.5953(±0.182) | 0.6436(±0.109) | 0.7262(±0.110) | **0.9722(±0.000)** |
| Mice | 0.3122(±0.024) | 0.3020(±0.014) | 0.3136(±0.014) | 0.3140(±0.030) | 0.2983(±0.041) | 0.2900(±0.019) | **0.3490(±0.008)** |
| COIL20 | 0.5593(±0.065) | 0.5126(±0.027) | 0.5813(±0.085) | 0.5113(±0.030) | **0.8329(±0.025)** | 0.4121(±0.018) | 0.8330(±0.002) |
| Mfeat | 0.6971(±0.077) | 0.6237(±0.063) | 0.6449(±0.091) | 0.6480(±0.044) | 0.8465(±0.106) | 0.8452(±0.091) | **0.9337(±0.001)** |
| Segment | 0.5783(±0.058) | 0.6281(±0.065) | 0.5806(±0.046) | 0.6441(±0.059) | 0.5799(±0.062) | 0.5289(±0.063) | **0.6965(±0.020)** |
| Digits | 0.7088(±0.069) | 0.7417(±0.092) | 0.7066(±0.066) | 0.7550(±0.064) | 0.8545(±0.092) | 0.8454(±0.086) | **0.8810(±0.001)** |
| Satimage | 0.6517(±0.069) | 0.6234(±0.062) | 0.6309(±0.082) | 0.6200(±0.064) | 0.6505(±0.063) | 0.6116(±0.040) | **0.7673(±0.020)** |
| MNIST10k | 0.5167(±0.039) | 0.4585(±0.047) | 0.4766(±0.039) | 0.4585(±0.047) | **0.7116(±0.087)** | 0.6573(±0.079) | 0.6307(±0.008) |
| PenDigits | 0.6883(±0.065) | 0.6905(±0.058) | 0.6429(±0.072) | 0.6905(±0.058) | 0.7920(±0.083) | 0.7392(±0.123) | **0.8086(±0.012)** |
| Shuttle | 0.4823(±0.143) | 0.5461(±0.165) | 0.7139(±0.134) | 0.4010(±0.083) | 0.5487(±0.122) | 0.3995(±0.036) | **0.8727(±0.021)** |
| Rank | 4.4 | 5.1 | 4.7 | 4.6 | 3.2 | 4.7 | **1.2** |

TABLE 4
The Average NMI (±Standard Deviation) of Seven Clustering Algorithms on 15 Real-world Datasets
(The Best Result of Each Dataset is Highlighted in Bold).

| Dataset | K-means | Ncut | FastESC | SAMSC | U-SPEC | GCSED | MeanCut |
|---|---|---|---|---|---|---|---|
| Iris | 0.5854(±0.263) | 0.5746(±0.257) | 0.5669(±0.258) | 0.5907(±0.263) | 0.4629(±0.299) | 0.5189(±0.241) | **0.7599(±0.014)** |
| Wine | 0.5912(±0.284) | 0.5528(±0.277) | 0.5355(±0.273) | 0.5551(±0.278) | 0.5632(±0.264) | 0.6105(±0.286) | **0.8495(±0.007)** |
| Seeds | 0.5330(±0.239) | 0.4856(±0.233) | 0.4625(±0.222) | 0.4917(±0.234) | 0.4845(±0.222) | 0.5469(±0.243) | **0.7606(±0.000)** |
| Dermatology | 0.8123(±0.075) | 0.8310(±0.072) | 0.7908(±0.058) | 0.8382(±0.068) | 0.8091(±0.069) | 0.8417(±0.046) | **0.9077(±0.003)** |
| Control | 0.7337(±0.026) | 0.6810(±0.049) | 0.6136(±0.052) | 0.7236(±0.033) | 0.8198(±0.047) | 0.7144(±0.021) | **0.8217(±0.000)** |
| Breast-Cancer | 0.5266(±0.246) | 0.4434(±0.230) | 0.5070(±0.236) | 0.4414(±0.229) | 0.3186(±0.274) | 0.4076(±0.277) | **0.8178(±0.000)** |
| Mice | 0.2882(±0.037) | 0.2701(±0.021) | 0.2843(±0.033) | 0.2733(±0.018) | 0.2599(±0.053) | 0.2600(±0.058) | **0.5978(±0.002)** |
| COIL20 | 0.7519(±0.029) | 0.6895(±0.015) | 0.7431(±0.038) | 0.6966(±0.019) | **0.9335(±0.011)** | 0.7342(±0.022) | 0.9169(±0.002) |
| Mfeat | 0.6880(±0.045) | 0.6805(±0.029) | 0.6036(±0.053) | 0.6892(±0.019) | 0.8398(±0.030) | **0.8764(±0.021)** | 0.8523(±0.001) |
| Segment | 0.6204(±0.026) | 0.6357(±0.018) | 0.6324(±0.019) | 0.6439(±0.017) | 0.6393(±0.022) | 0.6097(±0.050) | **0.6884(±0.007)** |
| Digits | 0.6470(±0.050) | 0.6648(±0.048) | 0.6241(±0.046) | 0.6633(±0.049) | 0.8301(±0.043) | **0.8845(±0.029)** | 0.8209(±0.003) |
| Satimage | 0.5542(±0.046) | 0.5133(±0.091) | 0.4846(±0.089) | 0.5150(±0.044) | 0.6121(±0.041) | 0.0009(±0.004) | **0.6523(±0.013)** |
| MNIST10k | 0.4275(±0.027) | 0.4455(±0.025) | 0.3809(±0.031) | 0.4458(±0.025) | 0.6760(±0.021) | **0.7086(±0.025)** | 0.5695(±0.003) |
| PenDigits | 0.6787(±0.040) | 0.6733(±0.054) | 0.6480(±0.065) | 0.6733(±0.054) | 0.8209(±0.041) | 0.7902(±0.069) | **0.8366(±0.003)** |
| Shuttle | 0.4202(±0.051) | 0.5291(±0.155) | 0.3906(±0.189) | 0.3843(±0.028) | 0.5305(±0.093) | 0.2908(±0.044) | **0.5361(±0.057)** |
| Rank | 3.9 | 4.9 | 5.8 | 4.2 | 3.7 | 4.1 | **1.4** |

TABLE 5
The Average ARI (±Standard Deviation) of Seven Clustering Algorithms on 15 Real-world Datasets
(The Best Result of Each Dataset is Highlighted in Bold).

| Dataset | K-means | Ncut | FastESC | SAMSC | U-SPEC | GCSED | MeanCut |
|---|---|---|---|---|---|---|---|
| Iris | 0.4826(±0.236) | 0.4755(±0.228) | 0.4674(±0.223) | 0.4929(±0.232) | 0.3567(±0.250) | 0.3543(±0.175) | **0.7574(±0.018)** |
| Wine | 0.5435(±0.286) | 0.4961(±0.284) | 0.4957(±0.286) | 0.5091(±0.283) | 0.5267(±0.266) | 0.5561(±0.293) | **0.8699(±0.012)** |
| Seeds | 0.4664(±0.225) | 0.4181(±0.237) | 0.3879(±0.226) | 0.4256(±0.238) | 0.4544(±0.225) | 0.4873(±0.223) | **0.7826(±0.005)** |
| Dermatology | 0.6628(±0.146) | 0.7147(±0.138) | 0.6340(±0.095) | 0.7335(±0.121) | 0.6464(±0.122) | 0.7856(±0.110) | **0.8486(±0.002)** |
| Control | 0.5721(±0.034) | 0.5052(±0.061) | 0.4112(±0.083) | 0.5643(±0.053) | **0.6630(±0.093)** | 0.4955(±0.032) | 0.6436(±0.000) |
| Breast-Cancer | 0.6012(±0.302) | 0.3949(±0.255) | 0.6192(±0.281) | 0.3910(±0.255) | 0.2494(±0.269) | 0.3822(±0.298) | **0.8910(±0.000)** |
| Mice | 0.1560(±0.018) | 0.1390(±0.017) | 0.1486(±0.022) | 0.1445(±0.023) | 0.1082(±0.034) | 0.0691(±0.025) | **0.2415(±0.027)** |
| COIL20 | 0.5181(±0.062) | 0.3128(±0.033) | 0.4996(±0.091) | 0.3056(±0.025) | 0.8003(±0.014) | 0.3544(±0.057) | **0.8136(±0.003)** |
| Mfeat | 0.6220(±0.085) | 0.5327(±0.081) | 0.5377(±0.080) | 0.5764(±0.029) | 0.8208(±0.094) | 0.8000(±0.091) | **0.9044(±0.001)** |
| Segment | 0.4676(±0.047) | 0.4945(±0.042) | 0.4614(±0.033) | 0.5054(±0.042) | 0.4662(±0.055) | 0.3457(±0.085) | **0.5895(±0.009)** |
| Digits | 0.5984(±0.078) | 0.6133(±0.177) | 0.5860(±0.083) | 0.6454(±0.079) | 0.8071(±0.138) | 0.7983(±0.095) | **0.8561(±0.003)** |
| Satimage | 0.5000(±0.050) | 0.4424(±0.073) | 0.4311(±0.100) | 0.4400(±0.073) | 0.5116(±0.062) | 0.4210(±0.086) | **0.6465(±0.026)** |
| MNIST10k | 0.3469(±0.023) | 0.2943(±0.024) | 0.3002(±0.030) | 0.2942(±0.024) | **0.6216(±0.064)** | 0.5572(±0.068) | 0.5041(±0.004) |
| PenDigits | 0.5605(±0.068) | 0.5570(±0.067) | 0.5093(±0.090) | 0.5570(±0.067) | 0.6931(±0.118) | 0.6242(±0.143) | **0.7647(±0.004)** |
| Shuttle | 0.2338(±0.078) | 0.3597(±0.211) | 0.2939(±0.220) | 0.1614(±0.035) | 0.3116(±0.157) | 0.1287(±0.065) | **0.5880(±0.035)** |
| Rank | 3.7 | 4.8 | 5.5 | 4.6 | 3.5 | 4.6 | **1.2** |

**Datasets:** The datasets in Table 2, include three image datasets (COIL20 [25], Digits, MNIST10k [26]), and 12 UCI datasets (Iris, Wine, Seeds, Dermatology, Control, Breast Cancer, Mice, Mfeat, Segment, Satimage, PenDigits, Shuttle)

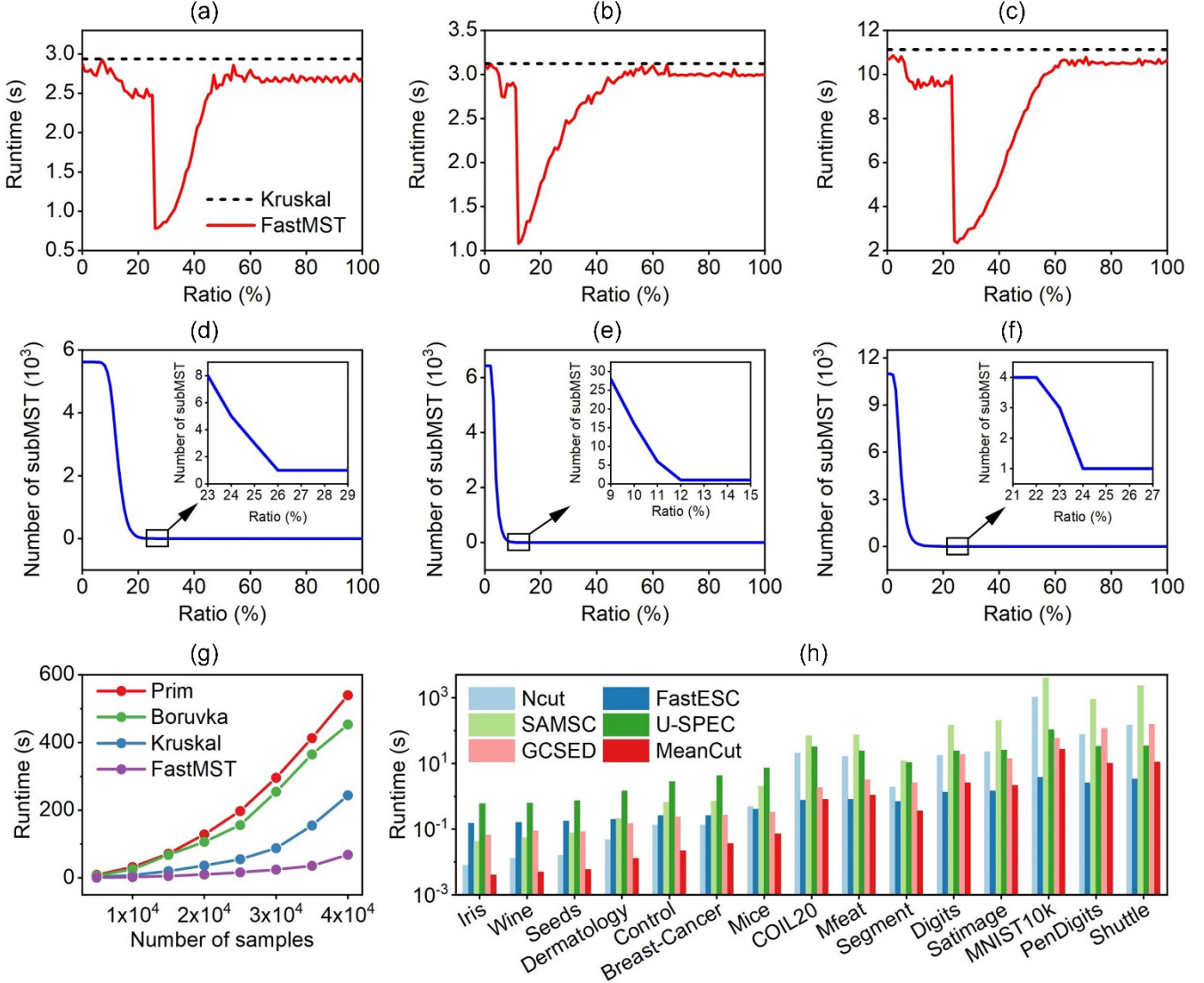

Fig. 6. Time efficiency analysis for MeanCut. (a)-(c) Runtimes and (d)-(f) number of subMST of FastMST under different *ratio* on Digits, Satimage and PenDigits respectively. (g) Runtimes of five MST algorithms on 2D synthetic datasets with different number of samples. (h) Runtimes of six graph clustering algorithms on 15 real-world datasets.

[27]. COIL20 consists of 32×32 grey-scale images of 20 objects. Digits and MNIST10k are two types of handwritten images with size of 8×8 and 28×28 respectively. The 12 UCI datasets contain diverse real-world instances with different dimensions of feature.

**Baselines:** We investigate six baselines, including two classical and four newly proposed spectral clustering algorithms. K-means is the most famous and widely used clustering algorithm [28], while Ncut is the representative spectral clustering method. Fast Explicit Spectral Clustering (FastESC) employs random Fourier features to represent data in kernel space explicitly, thereby reducing the complexity to solve eigenvectors [29]. Strength Adjusted Multilayer Spectral Clustering (SAMSC) is a multilayer community detection method based on a modified graph cut function [30]. Ultra-scalable SPEctral Clustering (U-SPEC) improves the computational time and memory by seeking K-nearest representatives and constructing a sparse affinity sub-matrix [31]. Graph Clustering with Spectral Embedding and Discretization (GCSED) adopts a new objective function by integrating embedding and rotation, and uses label propagation to accelerate the optimization [5].

**Parameter settings:** For the three image datasets, i.e., COIL20, Digits, MNIST10k, we convert the 2D image matrices to 1D vectors as the input. Before clustering, we adopt min-max normalization, i.e., $f_i = (f_i - f_{min})/(f_{max} - f_{min})$, to preprocess datasets, by scaling all features to $[0, 1]$. All the six baselines require the number of clusters. We set it to 1~7 on Iris, Wine, Seeds and Breast-Cancer, while from $k$-3 to $k$+3 on the remaining 11 datasets (here $k$ refers to the true number of classes). Other parameters are fixed as defaults. For MeanCut algorithm, we adopt a fully-connected graph, and calculate the initial edge weights as $w_{ij} = \exp(-\|v_i - v_j\|/2)$ using Laplacian kernel with $\sigma = 1$. We fix $ratio = 0.2$ for FastMST to keep a relatively stable time efficiency. The *percentile* threshold of DGF is tuned with an interval of 0.01 in the range of $[0.6, 1]$. $K$ of KNN is varied from 10 to 40 on small datasets, while from 20 to 60 on

MNIST10k, PenDigits and Shuttle, with an interval of 1. We traverse the specified parameter space, and select the top 20 clustering results of MeanCut for comparison.

**Evaluation metrics:** To evaluate the clustering accuracy quantitatively, we adopt three widely-used evaluation metrics, i.e., ACCuracy (ACC), Normalized Mutual Information (NMI), Adjusted Rand Index (ARI).

ACC refers to the accuracy rate of the clustering results compared with the true labels. We set the true label vector and the predicted label vector as $l = (l_1, l_2, ..., l_n) \in \mathbb{R}^n$ and $r = (r_1, r_2, ..., r_n) \in \mathbb{R}^n$ respectively, and the ACC can be defined as

$$\text{ACC} = \frac{\sum_{i=1}^n \delta(l_i, map(r_i))}{n} \tag{56}$$

where $\delta(\cdot)$ denotes an indicator function

$$\delta(x, y) = \begin{cases} 1 & \text{if } x = y \\ 0 & \text{otherwise} \end{cases} \tag{57}$$

$map(\cdot)$ is a mapping function that maps each predicted label to one of the true cluster labels. Commonly, the best mapping can be found by using the Kuhn-Munkres or Hungarian algorithms.

NMI measures the agreement of predict and true assignments by ignoring permutations. Its formulation is

$$\text{NMI} = \frac{\sum_{i=1}^{|S|} \sum_{j=1}^{|T|} |S_i \cap T_j| \log \frac{n|S_i \cap T_j|}{|S_i||T_j|}}{\sqrt{\left(\sum_{i=1}^{|S|} |S_i| \log \frac{|S_i|}{n}\right) \left(\sum_{j=1}^{|T|} |T_j| \log \frac{|T_j|}{n}\right)}} \tag{58}$$

where $S_i$ denote the predicted point set of the $i$th cluster, while $T_j$ denotes the $j$th cluster of the true labels.

ARI is modified based on the Rand Index (RI) as

$$\text{ARI} = \frac{\text{RI} - E(\text{RI})}{\max(\text{RI}) - E(\text{RI})} \tag{59}$$

where $E(\text{RI})$ denotes the expected RI score.

**Results:** The accuracies reported by ACC, NMI and ARI are illustrated in Table 3-5. In general, MeanCut outperforms the baselines strongly both in clustering accuracy and robustness. It obtains the highest average ACC and ARI on 13 datasets, and 11 highest average NMI scores. Meanwhile, MeanCut has lower standard deviations than the baselines, which indicates that it can produce more stable outcomes and is more robust to the parameters. In terms of the average rank, MeanCut achieves 1.2 rank in ACC and ARI, and 1.4 in NMI, which has distinct advantage than the competitors. It demonstrates the adaptability and utility of MeanCut.

### 5.2 Time Efficiency Analysis

We analyze the time efficiency of FastMST and the overall MeanCut procedure. As we discussed in Section 4.3, the time complexity of FastMST is affected by $ratio$. We vary $ratio$ and measure the runtime of FastMST on Digits, Satimage, and PenDigits. In Fig. 6a-c, as $ratio$ varies, the runtime first decreases and then increases. Both ends of the curve approach the runtime of Kruskal algorithm. The runtime reaches the minimum when the number of sub-MST just decreases to 1 as $ratio$ increases in Fig. 6d-f. Because the number of edges that need to be sorted reaches the minimal. From the results, it can be found that the optimal $ratio$ is commonly around 0.2. We fix $ratio$ as 0.2 and compare the runtime of FastMST with four classical MST algorithms on 2D synthetic datasets containing different number of samples. Results in Fig. 6g indicates that FastMST is more efficient than the classical MST algorithms and has a slower growth rate of runtime as the data size increases.

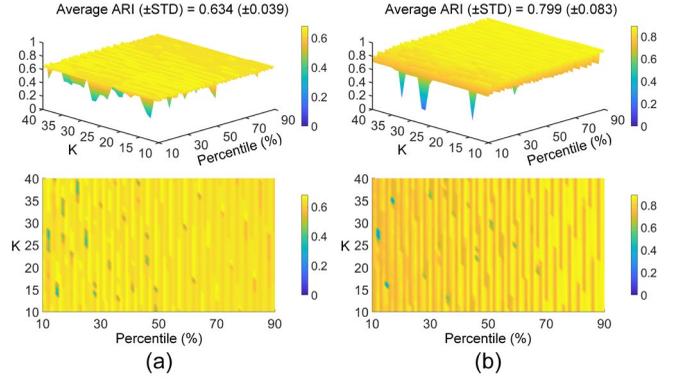

Fig. 7. Clustering accuracy indicated by ARI of MeanCut with respect to parameter $K$ and $percentile$ threshold on (a) Control and (b) Breast-Cancer datasets, where the two rows present the 3D and 2D views of ARI scores respectively.

We also testify the overall runtime of MeanCut, and compare with five spectral clustering baselines on the 15 real-world datasets. As illustrated in Fig. 6h, MeanCut has higher time efficiency than Ncut, SAMSC, U-SPEC, GCSED and runs faster than FastESC on the first 10 datasets. Moreover, MeanCut embraces parallel computing due to the nature of KNN-based calculation. It can be easily extended to parallel versions using GPGPU and distributed computing techniques such as Apache Spark, for performance acceleration.

### 5.3 Parameter Sensitivity Analysis

Parameter sensitivity analysis aims to measure the degree to which the clustering quality is affected by the variation of the parameters. Among the three parameters of MeanCut, $ratio$ in FastMST affects the time efficiency but not the clustering results, so we only analyze the sensitivity of $K$ of KNN and the $percentile$ threshold of DGF.

As shown in Fig. 7, we perform MeanCut on Control and Breast-Cancer datasets, and report the ARI scores by varying the two parameters. The $percentile$ threshold of DGF is adjusted with an interval of 0.01 in the range of [0.1, 0.9], and $K$ of KNN is tuned from 10 to 40 with an interval of 1. The clustering accuracy is stable in general, and most of ARIs fluctuate within a narrow range. In detail, the average ARI on Control is 0.634 with a small standard deviation of 0.039 (0.799 and 0.083 on Breast-Cancer respectively). It validates the robustness of MeanCut to the parameters.

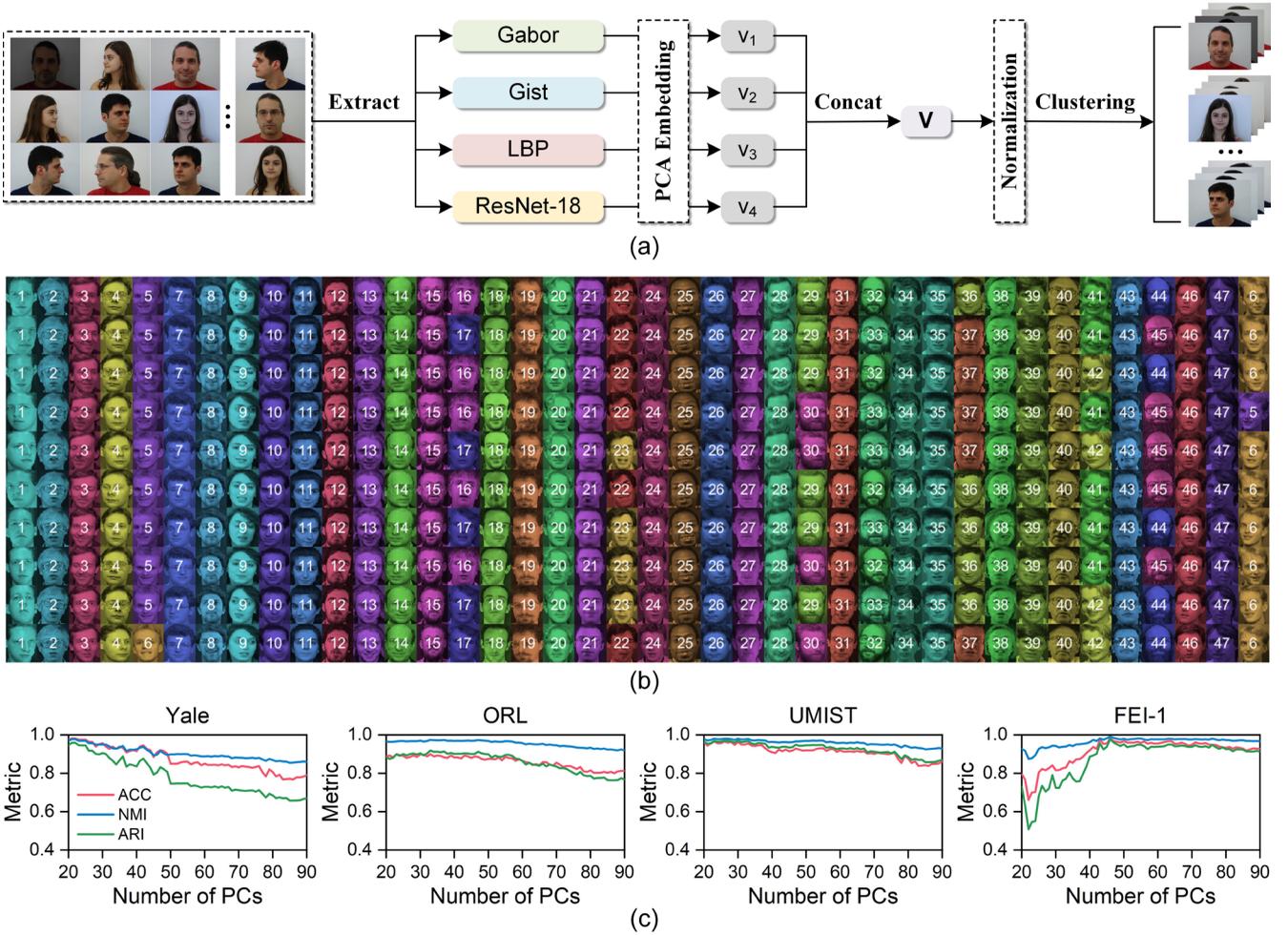

Fig. 8. Application of face recognition. (a) The workflow of unsupervised face clustering for MeanCut. (b) Pictorial representation of the best MeanCut cluster assignations on ORL dataset, where each column contains 10 images of one person and each color denotes a cluster. (c) The highest ACC, NMI and ARI of MeanCut on different PCA dimensions for the feature of All view.

### 5.4 Application of Face Recognition

To testify the adaptability in face recognition, we collect four face image datasets, and compare MeanCut with two versatile and six cutting-edge multi-view subspace clustering algorithms.

**Datasets:** Four selected face image datasets, i.e., Yale [32], ORL [33], UMIST [34], FEI-1 [35], contain human faces with different sexes, poses, views, lighting, accessories, and facial expressions. Yale has totally 165 face images with size of 64×64 from 15 distinct individuals. ORL contains 400 face images are greyscale with 112×92 pixels in size. They are collected from 40 persons and each person has 10 face images. UMIST consists of 564 images of 20 people, which have 220×220 pixels in 256 shades of grey. FEI-1 is comprised of 700 face images taken from 50 individuals and the images are in size of 480×640 pixels. Due to the large image size of FEI-1, they are resized to 64×64 before extracting the features.

**Baselines:** We investigate two versatile clustering algorithms, i.e., K-means and Ncut, and six cutting-edge multi-view subspace clustering algorithms that specialize in image classification. Co-Regularized multi-view spectral clustering (Co-Reg) uses the philosophy of co-regularization to make the clusters on different views agree with each other [36]. Robust Multi-view Subspace Clustering (RMSC) adopts a Markov chain to handle the possible noise explicitly in the transition probability matrices associated with different views [37]. Online Multi-View Clustering (OMVC) can deal with large-scale incomplete views by introducing dynamic weight setting to tackle the incompleteness of views and utilizing the Hessian matrices to accelerate gradient descent [38]. Pairwise/Centroid multi-view Low-Rank Sparse Subspace Clustering (PLRSSC/CLRSSC) learns a joint subspace representation by constructing affinity matrix shared among all views [39]. Consistent and Specific Multi-view Subspace Clustering (CSMSC) takes both consistency and specificity into account for subspace representation [40].

**Features:** We adopt four types of image features, including Gabor, Gist, LBP and a deep convolutional feature. Gabor can capture multi-scale and multi-orientation textural features with a good adaptability to the lighting. Here, the number of scales and orientations are set as 5 and 8 respectively. Gabor filter in size of 39×39 is used to convolve

TABLE 6
Performances of nine algorithms on four face image datasets.

| Dataset | Metric | Feature | K-means | Ncut | Co-Reg | RMSC | OMVC | PLRSSC | CLRSSC | CSMSC | MeanCut |
|---|---|---|---|---|---|---|---|---|---|---|---|
| Yale | ACC | Gabor | 0.7939 | 0.7212 | 0.6791 | 0.6994 | 0.6303 | 0.7636 | 0.7939 | 0.7455 | **0.8788** |
| | | Gist | 0.7091 | 0.7333 | 0.6897 | 0.7021 | 0.8121 | 0.7697 | 0.7879 | 0.6667 | **0.8667** |
| | | LBP | 0.5030 | 0.5697 | 0.5933 | 0.5681 | 0.6182 | 0.6727 | 0.6788 | 0.5697 | **0.7636** |
| | | ResNet-18 | 0.5152 | 0.6848 | 0.5609 | 0.5961 | 0.6061 | 0.6848 | 0.6545 | 0.5818 | **0.7091** |
| | | All view | 0.7333 | 0.7636 | 0.6312 | 0.6378 | 0.7091 | 0.7758 | 0.7879 | 0.8667 | **0.9818** |
| | NMI | Gabor | 0.7823 | 0.7402 | 0.7336 | 0.7181 | 0.6459 | 0.8053 | 0.8120 | 0.7713 | **0.8708** |
| | | Gist | 0.7873 | 0.7647 | 0.7589 | 0.7462 | 0.7959 | 0.8022 | 0.8166 | 0.6889 | **0.8502** |
| | | LBP | 0.5690 | 0.6035 | 0.6530 | 0.6058 | 0.6271 | 0.7032 | 0.7089 | 0.5925 | **0.7409** |
| | | ResNet-18 | 0.6326 | 0.7012 | 0.6139 | 0.6405 | 0.6617 | **0.7460** | 0.7142 | 0.6311 | 0.7150 |
| | | All view | 0.8111 | 0.7835 | 0.6886 | 0.6755 | 0.7249 | 0.8009 | 0.8109 | 0.8947 | **0.9772** |
| | ARI | Gabor | 0.5942 | 0.5396 | 0.5229 | 0.5303 | 0.4118 | 0.6470 | 0.6756 | 0.5966 | **0.7693** |
| | | Gist | 0.6009 | 0.5551 | 0.5651 | 0.5632 | 0.6377 | 0.6665 | 0.6742 | 0.4984 | **0.7221** |
| | | LBP | 0.2957 | 0.3245 | 0.4093 | 0.3702 | 0.3926 | 0.4924 | 0.5017 | 0.3626 | **0.5507** |
| | | ResNet-18 | 0.3441 | 0.4941 | 0.3773 | 0.4057 | 0.4077 | **0.5643** | 0.5005 | 0.4072 | 0.4945 |
| | | All view | 0.6193 | 0.5770 | 0.4502 | 0.4703 | 0.5370 | 0.6378 | 0.6712 | 0.7893 | **0.9594** |
| ORL | ACC | Gabor | 0.7950 | 0.7825 | 0.7344 | 0.7297 | 0.7400 | 0.8175 | 0.8250 | 0.7675 | **0.8800** |
| | | Gist | 0.7125 | 0.7850 | 0.7117 | 0.7246 | 0.7675 | 0.7900 | 0.7950 | 0.7350 | **0.8550** |
| | | LBP | 0.7350 | 0.7500 | 0.7440 | 0.7567 | 0.7275 | 0.8425 | **0.8600** | 0.7525 | 0.8050 |
| | | ResNet-18 | 0.7525 | 0.7650 | 0.7610 | 0.8032 | 0.6975 | 0.8450 | 0.8575 | **0.8600** | 0.8250 |
| | | All view | 0.7600 | 0.8150 | 0.7664 | 0.7769 | 0.8325 | 0.8850 | 0.8650 | 0.8525 | **0.9000** |
| | NMI | Gabor | 0.9059 | 0.8993 | 0.8733 | 0.8706 | 0.8598 | 0.9160 | 0.9156 | 0.9186 | **0.9373** |
| | | Gist | 0.8698 | 0.8981 | 0.8613 | 0.8686 | 0.8850 | 0.8971 | 0.8957 | 0.8692 | **0.9286** |
| | | LBP | 0.8866 | 0.9055 | 0.8845 | 0.8915 | 0.8694 | 0.9319 | **0.9413** | 0.8958 | 0.9316 |
| | | ResNet-18 | 0.9057 | 0.9163 | 0.8950 | 0.9097 | 0.8252 | 0.9264 | 0.9289 | **0.9355** | 0.9197 |
| | | All view | 0.9173 | 0.9402 | 0.8932 | 0.8972 | 0.9231 | 0.9562 | 0.9410 | 0.9511 | **0.9742** |
| | ARI | Gabor | 0.7221 | 0.6926 | 0.6539 | 0.6379 | 0.6149 | 0.7595 | 0.7646 | 0.7425 | **0.8160** |
| | | Gist | 0.6198 | 0.7025 | 0.6174 | 0.6337 | 0.6641 | 0.7096 | 0.6990 | 0.6402 | **0.7827** |
| | | LBP | 0.6474 | 0.6849 | 0.6763 | 0.6869 | 0.6263 | 0.7982 | **0.8269** | 0.6884 | 0.7832 |
| | | ResNet-18 | 0.7063 | 0.7100 | 0.7048 | 0.7359 | 0.5458 | 0.7881 | **0.8015** | 0.7975 | 0.7633 |
| | | All view | 0.7184 | 0.7704 | 0.7004 | 0.7203 | 0.7613 | 0.8636 | 0.8232 | 0.8124 | **0.9188** |
| UMIST | ACC | Gabor | 0.5061 | 0.4974 | 0.4956 | 0.4962 | 0.4661 | 0.5339 | 0.5530 | 0.4696 | **0.8209** |
| | | Gist | 0.5130 | 0.5200 | 0.5169 | 0.5142 | 0.4417 | 0.5687 | 0.5913 | 0.5287 | **0.8957** |
| | | LBP | 0.6713 | 0.6609 | 0.6688 | 0.6555 | 0.6139 | 0.6991 | 0.6852 | 0.6017 | **0.9443** |
| | | ResNet-18 | 0.6400 | 0.6557 | 0.5777 | 0.5904 | 0.5478 | 0.6643 | 0.6591 | 0.5930 | **0.8226** |
| | | All view | 0.6261 | 0.6017 | 0.5557 | 0.5872 | 0.5443 | 0.5548 | 0.5635 | 0.4730 | **0.9652** |
| | NMI | Gabor | 0.7530 | 0.7642 | 0.7295 | 0.6867 | 0.6098 | 0.7857 | 0.7796 | 0.6674 | **0.9159** |
| | | Gist | 0.7485 | 0.7566 | 0.6944 | 0.6917 | 0.5698 | 0.7692 | 0.7852 | 0.6711 | **0.9427** |
| | | LBP | 0.8339 | 0.8389 | 0.7961 | 0.8154 | 0.7512 | 0.8555 | 0.8532 | 0.7494 | **0.9702** |
| | | ResNet-18 | 0.8143 | 0.8243 | 0.7567 | 0.7731 | 0.7063 | 0.8316 | 0.8397 | 0.7623 | **0.9049** |
| | | All view | 0.8281 | 0.8345 | 0.7637 | 0.7451 | 0.7064 | 0.7916 | 0.7814 | 0.7189 | **0.9805** |
| | ARI | Gabor | 0.4197 | 0.4268 | 0.4087 | 0.3826 | 0.3104 | 0.5019 | 0.5044 | 0.3483 | **0.8333** |
| | | Gist | 0.4528 | 0.4431 | 0.4122 | 0.4041 | 0.2873 | 0.4955 | 0.5044 | 0.4053 | **0.9118** |
| | | LBP | 0.6095 | 0.6024 | 0.5872 | 0.5930 | 0.5006 | 0.6441 | 0.6297 | 0.5114 | **0.9499** |
| | | ResNet-18 | 0.5684 | 0.5740 | 0.4909 | 0.5144 | 0.4251 | 0.6178 | 0.6152 | 0.4976 | **0.8391** |
| | | All view | 0.6056 | 0.5514 | 0.4712 | 0.4943 | 0.4214 | 0.5023 | 0.4958 | 0.3946 | **0.9701** |
| FEI-1 | ACC | Gabor | 0.6286 | 0.6871 | 0.6459 | 0.6675 | 0.6471 | 0.7529 | 0.7514 | 0.7157 | **0.8671** |
| | | Gist | 0.5471 | 0.5886 | 0.6096 | 0.6060 | 0.6200 | 0.6929 | 0.7057 | 0.5271 | **0.7286** |
| | | LBP | 0.5914 | 0.6171 | 0.6604 | 0.6758 | 0.5586 | 0.7300 | 0.7257 | 0.6471 | **0.7829** |
| | | ResNet-18 | 0.7300 | 0.8057 | 0.7815 | 0.7847 | 0.6171 | 0.9029 | **0.9186** | 0.8114 | 0.8414 |
| | | All view | 0.7557 | 0.7671 | 0.8088 | 0.7936 | 0.7457 | 0.7471 | 0.7557 | 0.8129 | **0.9814** |
| | NMI | Gabor | 0.8326 | 0.8323 | 0.8120 | 0.8225 | 0.8173 | 0.8698 | 0.8646 | 0.8554 | **0.8973** |
| | | Gist | 0.7758 | 0.7872 | 0.7951 | 0.7934 | 0.8071 | 0.8309 | 0.8354 | 0.7475 | **0.8592** |
| | | LBP | 0.7637 | 0.7766 | 0.8081 | 0.8153 | 0.7167 | 0.8431 | 0.8320 | 0.8106 | **0.8667** |
| | | ResNet-18 | 0.8941 | 0.9054 | 0.8909 | 0.8873 | 0.7793 | 0.9426 | **0.9512** | 0.9060 | 0.9098 |
| | | All view | 0.9085 | 0.8855 | 0.9016 | 0.8998 | 0.8659 | 0.8677 | 0.8652 | 0.9137 | **0.9915** |
| | ARI | Gabor | 0.5503 | 0.5504 | 0.5313 | 0.5566 | 0.5257 | 0.6621 | 0.6477 | 0.6378 | **0.7572** |
| | | Gist | 0.3865 | 0.4650 | 0.4967 | 0.4897 | 0.4998 | 0.5735 | 0.5809 | 0.4055 | **0.6241** |
| | | LBP | 0.4169 | 0.4478 | 0.5545 | 0.5671 | 0.4089 | 0.6319 | 0.6128 | 0.5640 | **0.6605** |
| | | ResNet-18 | 0.6373 | 0.7258 | 0.7117 | 0.7023 | 0.4764 | 0.8596 | **0.8697** | 0.7517 | 0.7497 |
| | | All view | 0.6894 | 0.6691 | 0.7485 | 0.7354 | 0.6407 | 0.6527 | 0.6573 | 0.7661 | **0.9814** |

with 15×15 image blocks that are downsampled from the raw images. Gist is a global feature that models the shape of the scene, and it would generate a vector of 512 dimensions for each image. Local Binary Pattern (LBP) encodes the small-scale texture information. We let LBP work in a 4×4 pixel and it outputs a vector with 1239 dimensions. The ResNet-18 model is utilized to extract the convolution feature representation [41]. It has 18 layers and is pre-trained on more than a million images from the ImageNet database. The network requires an image input size of 227×227 and outputs a 1,000-dimensional vector. We integrate all these four features as a whole "All view" feature.

**Parameter settings:** For K-means, Ncut and MeanCut, the extracted features are embedded to the same lower dimensions by PCA before clustering, and then are processed by min-max normalization (in Fig. 8a). The number of Principal Components (PCs) is set 20~90, while the other multi-view clustering algorithms are fed with the extracted features directly. All the eight baselines require the input of cluster numbers, which are varied from $k$-5 to $k$+5 ($k$ refers to the true number of classes). For the six multi-view subspace clustering algorithms, the numbers of views are specified based on the number of input features. Other parameters are fixed as the defaults. MeanCut constructs a fully-connected graph and using Laplacian kernel function to measure the edge weights. We set *ratio* = 0.2 for FastMST and tune the *percentile* threshold of DGF in the range of [0.1, 0.8], and $K$ of KNN is varied from 2 to 20. The best results of each algorithm are compared.

**Results:** We leverage ACC, NMI and ARI to evaluate the clustering accuracy and report the best results in Table 6. The results show that MeanCut outperforms the baselines, and even surpassing by large margins on partial features. It obtains the highest ACC, NMI and ARI scores on the Gabor, Gist and All view features of all datasets. Taking All view as an example, the ARI scores of MeanCut are larger 0.1701, 0.0552, 0.3645 and 0.2153 than the second highest ARI scores on Yale, ORL, UMIST and FEI-1 respectively. Meanwhile, feature fusion is useful for MeanCut and improves its performance to a large extent. All of the three evaluation metrics on All view surpass 0.9 and are even close to 1 on Yale, UMIST and FEI-1. The ARI scores obtained on All view are larger 0.1901, 0.1028, 0.0202 and 0.2242 than the highest ARI scores on the single features. Fig. 8b presents the best MeanCut result based on the feature of All view on ORL dataset, where 31 of the 40 individuals have been recognized correctly. As PCA dimensionality increases, the clustering accuracy using the All view feature is relatively stable on ORL and UMIST as shown in Fig. 8c. While MeanCut needs less PCs on Yale, and more PCs on FEI-1 to obtain promising accuracy. Because Yale is a simpler dataset with small image size, while FEI-1 contains more human face images with complex lighting and views.

## 6 CONCLUSION

In this paper, we propose a novel graph clustering algorithm MeanCut based on greedy optimization. Specifically, the edge weights of graph are measured with path-based similarity, rather than the Euclidean-based metric. It enhances the intra-cluster associations and favors the identification of arbitrary shaped clusters. A pointwise greedy optimization strategy using degree descent criterion is exploited to minimize the MeanCut function. It evades the information loss caused by spectral embedding and discretization. Furthermly, FastMST is proposed to reduce the computational complexity of path-based similarity calculation, and a density gradient factor DGF is designed for the separation of weakly connected clusters. Extensive experiments have been conducted on multiple real-world and synthetic datasets. Results testify the adaptability of MeanCut, and demonstrate its accuracy, robustness, and time efficiency.